\documentclass[final,authoryear,5p,times,twocolumn]{elsarticle}

\usepackage{amsmath}
\usepackage{amssymb} 

\usepackage{tikz}
\usetikzlibrary{arrows,matrix,positioning,patterns}
\usetikzlibrary{calc}
\usepackage{pgfplots} 
\usetikzlibrary{external}
\usetikzlibrary{angles}
\usepackage{hyperref}
\usetikzlibrary{plotmarks}

\newdefinition{rmk}{Remark}
\newproof{pf}{Proof}
\newproof{pot}{Proof of Theorem \ref{thm2}}

\usepackage[ruled, vlined, linesnumbered]{algorithm2e}

\usepackage{amsthm}

\newtheorem{remark}{Remark}

\newtheorem{problem}{Problem}

\usepackage{amsmath}
\usepackage{amssymb} 

\usepackage{eurosym}
\usepackage{colortbl}

\usetikzlibrary{arrows.meta,calc,decorations.markings,math,arrows.meta}

\usepackage[labelformat=simple]{subcaption}
\DeclareCaptionLabelSeparator{periodspace}{.\quad}
\usepackage{color,framed}
\usepackage[utf8]{inputenc} 
\usepackage{tabularx}

\usepackage{rotating}
\usepackage{graphicx}
\usepackage{lscape}
\allowdisplaybreaks 
\usepackage{longtable}

\usepackage[utf8]{inputenc}
\usepackage{pgfplots}
\usepgfplotslibrary{groupplots} 
\pgfplotsset{compat=1.3}

\makeatletter
\def\ps@pprintTitle{%
 \let\@oddhead\@empty
 \let\@evenhead\@empty
 \def\@oddfoot{}%
 \let\@evenfoot\@oddfoot}
\makeatother

\begin{document}

\begin{frontmatter}
%
%
\title{Smoothing of Headland Path Edges and Headland-to-Mainfield Lane Transitions\\ Based on a Spatial Domain Transformation and Linear Programming}
\author{Mogens Plessen\corref{cor1}}
\cortext[cor1]{MP is with Findklein GmbH, Switzerland, \texttt{mgplessen@gmail.com}}

\begin{abstract}
Within the context of in-field path planning, and under the assumption of nonholonomic vehicle models, this paper addresses two tasks: smoothing of headland path edges and smoothing of headland-to-mainfield lane transitions. Both tasks are solved by a two-step hierarchical algorithm. The first step differs for the two tasks generating either a piecewise-affine or a Dubins reference path. The second step leverages a transformation of vehicle dynamics from the time domain into the spatial domain and linear programming. Benefits, such as a hyperparameter-free objective function and spatial constraints useful for area coverage gaps avoidance and precision path planning, are discussed. The method, which is a deterministic optimisation-based method, is evaluated on 5 real-world fields solving 19 instances of the first task and 84 instances of the second task.
\end{abstract} 
\end{frontmatter}

\section{Introduction\label{sec_intro}}

\begin{table}
\centering
\begin{tabular}{|ll|}
\hline
\multicolumn{2}{|c|}{NOMENCLATURE TABLE}\\
\multicolumn{2}{|l|}{Abbreviations}\\
LP & Linear Programming. \\
PWA & Piecewise-Affine. \\
UTM & Universal Transverse Mercator coordinate system. \\[3pt]
\multicolumn{2}{|l|}{Symbols}\\
$D_s$ & Spatial discretisation spacing, (m).\\
$R_\text{Dubins}$ & Radius of Dubins path, (m).\\
$e_\psi$ & Path-aligned relative heading, ($^{\circ}$).\\
$l$ & Wheelbase of vehicle, (m).\\
$\bar{n}_\text{cstrts}$ & Average number of inequality constraints, (-).\\
$\bar{n}_\text{u}$ & Average number of decision variables, (-).\\
$(s,e_y)$ & Path-aligned position coordinates, (m).\\
$v$ & Velocity, (ms$^{-1}$).\\
$w$ & Operating width (inter-lane distance), (m).\\
$(x,y)$ & Global position coordinates, (m).\\[3pt]
\multicolumn{2}{|l|}{Greek Symbols}\\
$\delta$ & Steering angle of vehicle, ($^{\circ}$).\\
$\psi$ & Global absolute heading, ($^{\circ}$).\\
$\rho_s$ & Radius of curvature, (m).\\
$\bar{\tau}$ & Average solution runtime, (s).\\
\hline
\end{tabular}
\end{table}

Logistics play a large role in agriculture. For arable farming and the cultivation of crops it can be differentiated between in-field and inter-field logistics (\cite{plessen2019coupling}). For the former, efficient path planning is of interest subject to three main considerations: (i) minimisation of tractor traces within the field in order to maximise growable field area, (ii) minimisation of area coverage pathlength in order to minimise fuel consumption and time spent in the field, and (iii) avoiding area coverage gaps so that the entire field is covered, for example during a field-run spraying herbicide or pesticide applications, while simultaneously accounting for a given operating width of the machinery and accounting for its typically nonholonomic vehicle dynamics, state and actuator constraints.

Within this context the motivation and contribution of this paper is to present a method for smoothing of headland path edges and headland-to-mainfield lane transitions based on a spatial domain transformation and linear programming. As will be motivated below, the spatial domain transformation permits the formulation of spatial constraints useful for precision path planning in agriculture. A linear programming framework with a hyperparameter-free objective function can incorporate those constraints together with linearised and discretised nonholonomic vehicle dynamics in a disciplined manner to produce deterministic precision paths and fast solution times.

This research topic is of interest to both autonomous robot applications as well as for manual driving by providing a guidance reference to a human driver. Relevance is underlined, first, by the ubiquity of headland path and headland-to-mainfield lane transitions for every field coverage, and, second, by the nonlinear nature of the problem for nonholonomic vehicle dynamics.

Thus, given an optimised sequence of mainfield lane traversals for field coverage, such as produced in \cite{plessen2019optimal} and \cite{plessen2018partial}, this paper provides a method for path smoothing. See also \cite{hoffmann2024optimal}, \cite{mier2023fields2cover}, \cite{khan2016coverage} for more high-level area coverage techniques. This paper differs from \cite{plessen2017reference}, where an \emph{online} closed-loop tracking controller based on model predictive control was presented. In this work, the focus is on \emph{offline} generation of precision paths, which might then afterwards be tracked online by the approach from aforementioned paper or an alternative closed-loop tracking controller.

A widespread approach to reference path generation, and in particular headland turn trajectory generation, is to build methods around specific path planning \emph{patterns}. In \cite{trendafilov2022comparative}, 5 types of T-turns are considered. In \cite{wang2018adaptive} adaptive turning based on a switchback-pattern and dynamic circleback-pattern is discussed. In \cite{sabelhaus2013using}, 9 different symmetric lane-to-lane transitions are analysed, including Omega-turn, U-turn, gap-turn and fishtail-turn. Similarly, \cite{backman2015smooth} accounted for maximum steering rate and the maximum acceleration of the vehicle. In \cite{bulgakov2019theoretical}, for a trailed asymmetric swath reaper–tractor aggregate a pear-shaped loop-turn and a U-turn are discussed. In \cite{boryga2020application}, 4 types of patterns are discussed, whereby each is composed of two transition curves. In \cite{peng2023optimization}, an optimisation‐based method for turning in constrained headland spaces, such as orchards, is presented, using 4 types of turns (U-turn, Omega-turn, switchback-turn and circleback-turn) for generation of a first reference path. \cite{gao2023efficient} used Bezier curves for a headland-turning navigation system for a safflower picking robot. Finally, see \cite{ha2018development} for an example of a hardware modification to facilitate headland turning.

Once a smooth path plan is computed, for autonomous robots, a plethora of techniques for online tracking can be applied, see e.g. \cite{wang2023modelling}, \cite{ettefagh2018laguerre}, \cite{he2023adaptive}, \cite{wang2024accurate}, \cite{sun2024novel}, \cite{liu2024fuzzy}, \cite{eski2019control}, \cite{jing2021path}, \cite{yin2020trajectory}. For a recent overview of control algorithms of autonomous all-terrain vehicles in agriculture see \cite{etezadi2024comprehensive}.

The research gap and motivation for this paper is as follows. First, the focus of aforementioned pattern-based references is on lane-to-lane transitions. Second, an issue for pattern-based path planning, in particular based on straights, arcs and clothoid segments, is that resulting paths are typically operating at actuation limits, such as maximum curvature and maximum steering rate. For general headland path smoothing (in contrast to aforementioned lane-to-lane transitions), this is not sufficient where other constraints, such as area coverage gaps, must also be accounted for. Third, the proposal of a spatial domain transformation in order to be able to formulate optimisation problems with spatial constraints, which is desirable for precision agriculture applications, is missing in the literature.

The focus of this paper is on path planning instead of path tracking. The allure of a numerical optimisation approach is that it allows incorporation of a variety of constraints in a disciplined manner and can handle the multivariate nature of the task. The advantage of transforming vehicle dynamics from the time into the spatial domain is that it allows to formulate spatial precision constraints. An early paper discussing benefits of transformations of dynamics from the time into the spatial domain is \cite{pfeiffer1987concept}. To the author's knowledge this is the first paper that proposes a spatial transformation of system dynamics to calculate precision path plans in agriculture.

The rest of the paper is organised as follows: problem formulation, modeling and proposed solution, numerical results and the conclusion are described in Sections \ref{sec_probformulation}-\ref{sec_conclusion}.

\section{Problem Formulation and Solution\label{sec_probformulation}}

\begin{figure*}
\centering
\includegraphics[width=7.8cm]{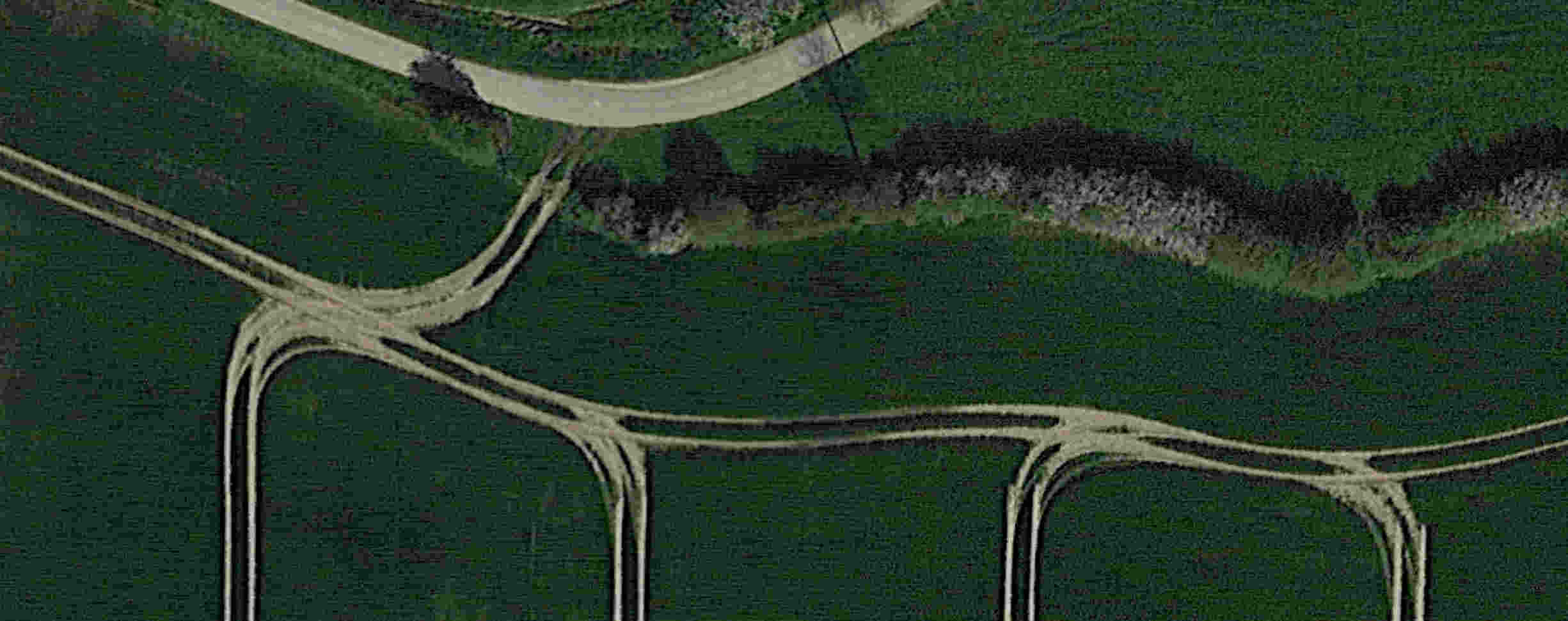}\hspace{1.5cm}
\begin{tikzpicture}
\draw [blue] plot [rounded corners=0.5cm] coordinates { (1,-0.1)(1,2)(3,2)(3,-0.1)};
\draw [blue] plot [rounded corners=0.5cm] coordinates { (-0.5,2)(4.5,2)};
%
%
\draw [draw=black,draw opacity=1, line width=10pt] plot [rounded corners=0.25cm] coordinates { (1,0.4)(1,0.8)};
%
\draw [draw=black,draw opacity=1, line width=1pt] plot [rounded corners=0.25cm] coordinates { (0,0.4)(2,0.4)};
\draw [draw=black,draw opacity=0.3, line width=26pt] plot [rounded corners=0.25cm] coordinates { (0,-0.1)(2,-0.1)};
\draw [black,-{Latex[scale=1.0]}] plot [rounded corners=0.25cm] coordinates { (1,0.8)(1,1.3)};
\node[color=black] (a) at (3.77, 1.01) {mainfield};
\node[color=black] (a) at (3.4, 0.66) {lane};
\node[color=black] (a) at (0.19, 0.87) {vehicle};
\node[color=black] (a) at (1.1, 2.3) {headland path};
\node[color=black] (a) at (-1.1, -0.25) {sprayed};
\node[color=black] (a) at (-1.1, -0.65) {area};
\draw[black] (-0.35,-0.22) -- (0.15,-0.15);
\node[color=black] (a) at (1, -0.33) {Q};
\node[color=black] (a) at (3, -0.33) {R};
\node[color=black] (a) at (-0.8, 2) {C};
\node[color=black] (a) at (4.8, 2) {D};
\end{tikzpicture}
\caption{\emph{Left}: Visualisation of real-world transitions between headland path and mainfield lanes. As illustrated, the transitions between headland path and mainfield lanes can be ``messy'' in the sense that these often cause an increased amount of compacted field areas due to crossings of tyre traces. \emph{Right}: Abstract visualisation with the definitions of headland path and mainfield lanes along which a vehicle (e.g., with spraying implement) might travel from location Q towards R or D.}
\label{fig_problFormulation}
\end{figure*}

\begin{figure}
\centering
\includegraphics[width=8.5cm]{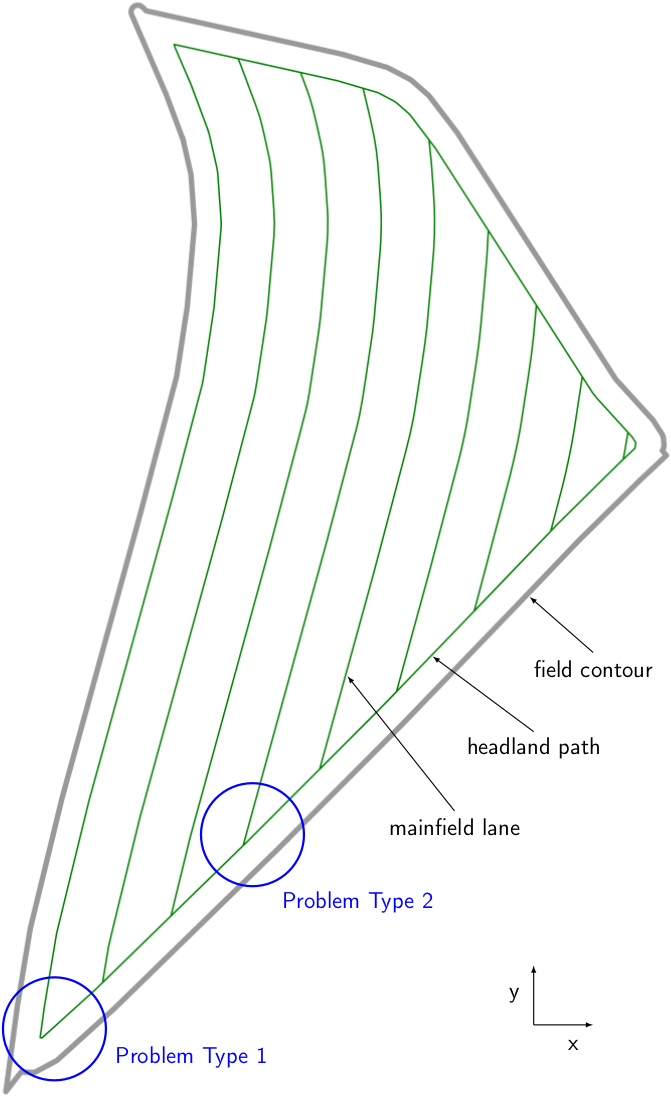}
\caption{Illustration of terminology and visualisation of Problem Type \ref{problem1} and \ref{problem2}.}
\label{fig_f0}
\end{figure}  

Figure \ref{fig_problFormulation} introduces basic terminology. In a first step, with the objective of generating an area covering path for a given field in mind, a lane-grid consisting of a headland path and mainfield lanes is generated, see Fig. \ref{fig_f0}. Therefore, the field contour is eroded to generate the headland path, before mainfield lanes are fitted. Mainfield lanes can either be straight lanes or freeform lanes, for example, adapted in shape using a partial segment of the headland path as reference. See \cite{plessen2021freeform} for the generation of freeform lanes and a comparison to straight lanes for the minimisation of the total number of mainfield lanes. The inter-lane distance is selected as the machinery operating width denoted by $w>0$.

After generating a lane-grid and under the assumption that the field shall be covered by a nonholonomic vehicle, two problem types arise. Both evolve around the smoothing of edges to generate a path suitable for nonholonomic vehicle dynamics, either along the headland path or at the transitions between mainfield lanes and the headland path.

\begin{problem}\label{problem1}
Given a headland path, identify path segments with sharp corners or abrupt directional changes, and modify these segments to produce a smoothed headland path that (i) is compatible with nonholonomic vehicle dynamics and (ii) avoids any area coverage gaps.
\end{problem}

\begin{problem}\label{problem2}
Given headland-to-mainfield lane transitions, and vice versa mainfield lane-to-headland transitions, as part of a field covering path, correct these if necessary such that smoothed transitions are generated that (i) are compatible with nonholonomic vehicle dynamics and (ii) avoid any area coverage gaps.
\end{problem}

The proposal of an algorithm for the solution of Problems \ref{problem1} and \ref{problem2} is the topic of this paper.

The general approach to address both Problems \ref{problem1} and \ref{problem2} is a hierarchical two-step algorithm. The first step differs for the two problems and has as objective the generation of a reference path. The second step uses that reference path and leverages linear programming and a transformation of vehicle dynamics from the time into the spatial domain.The general approach to address both Problems \ref{problem1} and \ref{problem2} is a hierarchical two-step algorithm. The first step differs for the two problems and has as objective the generation of a reference path. The second step uses that reference path and leverages linear programming and a transformation of vehicle dynamics from the time into the spatial domain.

\begin{figure}[ht!]
\centering
\begin{tikzpicture}
\coordinate (R) at (3,3.7);
\coordinate (F) at ($ (R) + 3*({cos(30)},{sin(30)})$);
\coordinate (C) at ($ (R) + 0*({cos(30)},{sin(30)})$);
\def\psis{18} 
\def \wheelL {0.5} 
\def \lineRhoSTop {3}
\def \lineRhoSBottom {2.5}
\draw[->] (0, 0) -- (C|-,0);
\draw[->] (0, 0) -- (0, |-C);
\node[color=black] (a) at ($ (C|-,0) + 0.25*({cos(0)},{sin(00)})$) {$x$}; 
\node[color=black] (a) at ($ (0, |-C) + 0.25*({cos(90)},{sin(90)})$) {$y$}; 
\draw[color=black] (R) -- (F);
\draw[color=black,|-|] ($ (C) + 1.15*({cos(-60)},{sin(-60)})$) -- ($ (F) + 1.15*({cos(-60)},{sin(-60)})$) node [midway, below] {$l$};
\draw[rounded corners,black,fill=black,fill opacity=0.2,rotate around={30:(R)}] ($ (R) - \wheelL*({cos(0)},{0.2*sin(90)})$) rectangle ($ (R) + \wheelL*({cos(0)},{0.2*sin(90)})$);
\draw[rounded corners,black,fill=black,fill opacity=0.2,rotate around={60:(F)}] ($ (F) - \wheelL*({cos(0)},{0.2*sin(90)})$) rectangle ($ (F) + \wheelL*({cos(0)},{0.2*sin(90)})$);
\fill (C) circle [radius=2pt];
\draw[color=black!50] (C) -- ($ (C) + 1.75*({cos(0)},{sin(0)})$);
\draw[->] ($ (C) + 0.75*({cos(0)},{sin(0)})$) arc (0:30:0.75);
\node[color=black] (a) at ($ (C) + 1*({cos(10)},{sin(10)})$) {$\psi$};
\draw[->] ($ (C) + 0.3*({cos(180)},{sin(180)})$) arc (180:320:0.3);
\node[color=black] (a) at ($ (C) + 0.6*({cos(320)},{sin(320)})$) {$\dot{\psi}$};
\draw[->,color=red!50,>=latex',ultra thick] (C) -- ($ (C) + 1.3*({cos(30)},{sin(30)})$);
\node[color=black] (a) at ($ (C) + ({0.9*cos(30)},{0.75*sin(90)}) + (-0.29,-0.06)$) {$v$};
\draw[color=black!50] (F) -- ($ (F) + 1.75*({cos(30)},{sin(30)})$);
\draw[color=black!50] (F) -- ($ (F) + 1.75*({cos(60)},{sin(60)})$);
\draw[->] ($ (F) + 1.15*({cos(30)},{sin(30)})$) arc (30:60:1.15);
\node[color=black] (a) at ($ (F) + 1.4*({cos(45)},{sin(45)})$) {$\delta$};
\fill[color=blue] ($ (C) + \lineRhoSTop*({cos(90+\psis)},{sin(90+\psis)})$) circle [radius=2pt];
\draw[color=blue,->] ($ (C) + \lineRhoSTop*({cos(90+\psis)},{sin(90+\psis)})$) -- ($ (C) + \lineRhoSBottom*({cos(-(90-\psis))},{sin(-(90-\psis))})$) node [pos=0.22, left=0.04] {$\rho_s(s)$} node [pos=0.99, left=0.05] {$s$};
\draw[color=black!50,->] ($ (C) + \lineRhoSTop*({cos(90+\psis)},{sin(90+\psis)}) + {\lineRhoSTop+\lineRhoSBottom}*({cos(-93)},{sin(-93)})$) arc (-93:-40:{\lineRhoSTop+\lineRhoSBottom});
\node[color=black!50] (a) at ($ (C) + 3.5*({cos(90+\psis)},{sin(90+\psis)}) + 6.5*({cos(-35)},{sin(-35)}) + (0.36,0.04)$) {path centerline};
\draw[color=blue,->] ($ (C) + \lineRhoSTop*({cos(90+\psis)},{sin(90+\psis)}) + 0.3*({cos(180)},{sin(180)})$) arc (180:330:0.3);
\node[color=blue] (a) at ($ (C) + \lineRhoSTop*({cos(90+\psis)},{sin(90+\psis)}) + 0.67*({cos(330)},{sin(330)})$) {$\dot{\psi}_s$};
\draw[-,color=black!50, dashed] ($ (C) + \lineRhoSBottom*({cos(-(90-\psis))},{sin(-(90-\psis))}) - 2*({cos(\psis)},{sin(\psis)}) $) -- ($ (C) + \lineRhoSBottom*({cos(-(90-\psis))},{sin(-(90-\psis))}) + \lineRhoSTop*({cos(\psis)},{sin(\psis)}) $) node [pos=0.95, right=0.33] {tangent};
\draw[->,color=blue,>=latex',ultra thick] ($ (C) + \lineRhoSBottom*({cos(-(90-\psis))},{sin(-(90-\psis))})$) -- ($ (C) + \lineRhoSBottom*({cos(-(90-\psis))},{sin(-(90-\psis))}) + 1.5*({cos(\psis)},{sin(\psis)}) $) node [pos=0.95, left=0.5] {$\dot{s}$};
\draw[-,color=black!50, dashed] ($ (C) - 2*({cos(\psis)},{sin(\psis)}) $) -- ($ (C) + 4*({cos(\psis)},{sin(\psis)}) $);
\draw[->, color=blue] ($ (C) + 1.4*({cos(\psis)},{sin(\psis)})$) arc (\psis:30:1.4);
\node[color=blue] (a) at ($ (C) + 1.65*({cos(22)},{sin(22)})$) {$e_\psi$};
\draw[->, color=blue] ($ (C) + 1.4*({cos(0)},{sin(0)})$) arc (0:\psis:1.4);
\node[color=blue] (a) at ($ (C) + 1.65*({cos(10)},{sin(10)})$) {$\psi_s$};
\draw[<-,color=blue] ($ (C) + 2*({cos(\psis)},{sin(\psis)}) $) -- ($ (C) + \lineRhoSBottom*({cos(-(90-\psis))},{sin(-(90-\psis))}) + 2*({cos(\psis)},{sin(\psis)}) $) node [pos=0.55, right=-0.05, color=blue] {$e_y$};
\end{tikzpicture}
\caption{The kinematic bicycle model including the representation of the path-aligned coordinate system, see \cite{rajamani2011vehicle,plessen2017spatial}. The centre-of-gravity (CoG) of the vehicle is assumed to be located at the rear-axle and at a distance of $l>0$ to the steerable front-axle. Control inputs are velocity $v>0$ and steering angle $\delta\in[-\delta_\text{max},\delta_\text{max}]$. In the global position coordinate system the CoG-position is denoted by $(x,y)\in\mathbb{R}^2$ and the vehicle heading is $\psi\in[0,2\pi]$. This vehicle state can likewise be expressed in a path-aligned coordinate system, where $e_\psi\in[-\pi,\pi]$ and $e_y\in\mathbb{R}$ represent relative heading and lateral deviations with respect to references along the path centreline coordinate $s>0$. The radius of curvature at that coordinate is denoted by $\rho_s(s)>0$.}
\label{fig_spatial_bicycle_mdl}
\end{figure} 

\section{Theory\label{sec_solnMethod}}

\subsection{Spatial Domain Transformation\label{subsec_spatial}}

To illustrate the spatial domain transformation and for simplicity, the method is illustrated for a kinematic bicycle model in Fig. \ref{fig_spatial_bicycle_mdl}. Readers are referred to \cite{plessen2017spatial} for the handling of more complex vehicle models. 

This section summarises the bicycle model using the notation of \cite{plessen2017spatial}. Consider a \textit{global} and a \textit{path-aligned} coordinate frame within the $(x,y)$- and $(s,e_y)$-plane, respectively, with path centreline coordinate $s\geq 0$ and lateral deviation $e_y\in\mathbb{R}$. Coordinates can be transformed by projecting point masses (vehicle's centre of gravity) to piecewise-affine path segments. As a detail, it can be differentiated between the path centreline coordinate $s\geq 0$, and actual path length coordinate $\eta\geq 0$, indicating the distance traveled by the vehicle. Note that $\eta \neq s$, unless the vehicle is traveling perfectly along the path centreline. A trajetory can then be defined as $z(s) = \begin{bmatrix} e_{\psi}(s),~e_{y}(s) \end{bmatrix}$, with heading deviation $e_{\psi}(s)\in[0,2\pi]$ at coordinate $s$. The equivalent trajectory in the $(x,y)$-plane is defined by $\mathcal{X}(s) = \begin{bmatrix} x(s),~y(s),~ \psi(s) \end{bmatrix}$. The classic nonlinear kinematic bicycle model (\cite{rajamani2011vehicle}) is
\begin{equation}
\begin{bmatrix} \dot{x},~\dot{y},~\dot{\psi} \end{bmatrix} = \begin{bmatrix} v\cos(\psi),~v\sin(\psi),~\frac{v}{l}\tan(\delta) \end{bmatrix},\label{eq_dotxypsi_nonlinkinbicmdl}
\end{equation}
assuming the centre of gravity to be located at the rear axle and $l$ denoting the wheelbase. The front-axle steering angle is $\delta$ and vehicle velocity is $v$. Let time and spatial derivatives be denoted by $\dot{x}=\frac{d x}{dt}$ and $x'=\frac{d x}{d s}$, respectively. The spatial equivalent of Eqn. \eqref{eq_dotxypsi_nonlinkinbicmdl} is derived as follows. Assuming $\dot{e}_\psi=\dot{\psi}-\dot{\psi}_s$ where $\psi_s\in[0,2\pi]$ the path heading, $\dot{e}_y=v \sin(e_\psi)$, $\dot{s}=\frac{\rho_s v \cos(e_\psi)}{\rho_s-e_y}$ where $\rho_s$ denotes the radius of curvature, $e_\psi'=\frac{\dot{e}_\psi}{\dot{s}}$ and $e_y' = \frac{\dot{e}_y}{\dot{s}}$, then
\begin{equation}
\begin{bmatrix} e_\psi' ,~ e_y' \end{bmatrix} \!=\! \begin{bmatrix} \frac{(\rho_s - e_y)\tan(\delta)}{\rho_s l \cos(e_\psi)} -\psi_s' ,~ \frac{\rho_s-e_y}{\rho_s} \tan(e_\psi)  \end{bmatrix}.\label{eq_nonlinmdl_steering}
\end{equation}
Importantly, Eqn. \eqref{eq_nonlinmdl_steering} is entirely independent of vehicle speed $v$. This is characteristic for kinematic models, but not the case for dynamic vehicle models (\cite{plessen2017spatial}). Furthermore, note that Eqn. \eqref{eq_nonlinmdl_steering} has one state dimension less than Eqn. \eqref{eq_dotxypsi_nonlinkinbicmdl}. To summarise, we define spatially-dependent state  and control vectors as $z=\begin{bmatrix}e_\psi ,~ e_y \end{bmatrix}$ and $u = \begin{bmatrix} \delta \end{bmatrix}$, respectively. Equation \eqref{eq_nonlinmdl_steering} is abbreviated by $z' = f(z,u)$. Let a discretisation grid along the path centreline be defined by $\{s_j\}_{j=0}^N = \{s_0,s_1,\dots,s_N\}$. For the remainder of this paper, index $j$ refers to the spatial discretisation grid. The discretisation grid can be uniformly spaced. Additional grid points can be added. For a given set of references $\{e_{\psi,j}^\text{ref}\}_{j=0}^N$, $\{e_{y,j}^\text{ref}\}_{j=0}^N$ and $\{u_j^\text{ref}\}_{j=0}^{N-1}$ and an initial state, the linearised and discretised (zero-order hold) state dynamics of Eqn.  \eqref{eq_nonlinmdl_steering} can be computed for $\{e_{\psi,j}\}_{j=0}^N$ and $\{e_{y,j}\}_{j=0}^N$ as a function of control inputs $\{\delta_j\}_{j=0}^{N-1}$. 


Continuous rate constraints on steering are,
\begin{equation}
\dot{\delta}_\text{min} \leq \dot{\delta} \leq \dot{\delta}_\text{max},
\label{eq_time_rateCstrts}
\end{equation}
whereby the bounds, $\dot{\delta}^\text{min}$ and $\dot{\delta}^\text{max}$, in general, are nonlinear functions of the vehicle's operating point and time-varying parameters. By applying the spatial coordinate transformation $\delta'=\frac{\dot{\delta}}{\dot{s}}$, a discretisation, a change of variables, and assuming the bounds to remain constant over the planning horizon, then

\begin{small}
\begin{align}
&  \frac{D_{s,j} \dot{\delta}_\text{min} (\rho_{s,j} - e_{y,j})}{ \rho_{s,j} \cos(e_{\psi,j}) v_j} \leq \delta_{j+1} - \delta_j \leq \frac{ D_{s,j} \dot{\delta}_\text{max} (\rho_{s,j} - e_{y,j}) }{ \rho_{s,j} \cos(e_{\psi,j}) v_j},\label{eq_deltad_rateCstrts}
\end{align}
\end{small}\normalsize
where $D_{s,j}=s_{j+1} - s_j$. Thus, the linear rate constraints in Eqn. \eqref{eq_time_rateCstrts} are rendered not only nonlinear, but also state-dependent and also velocity-dependent. This has two implications. First, to formulate linearly constrained optimisation problems, in general, the linearisation of Eqn. \eqref{eq_deltad_rateCstrts} is required. Depending on the quality of underlying reference trajectories, this may incur significant distortions. Second, while the discrete form of Eqn. \eqref{eq_time_rateCstrts} can always be guaranteed to be feasible (assuming a feasible initialisation), for Eqn. \eqref{eq_deltad_rateCstrts} this is not the case anymore. Consider two degrees of simplification of Eqn.  \eqref{eq_deltad_rateCstrts}. First, assume that the given reference trajectories permit small errors $e_{\psi,j}$ and $e_{y,j}$, and consequently approximate
$\frac{(\rho_{s,j} - e_{y,j})}{\rho_{s,j} \cos(e_{\psi,j})} \approx 1$, thereby rendering Eqn. \eqref{eq_deltad_rateCstrts} state-independent, but maintaining velocity-dependent bounds. Thus, steering rate constraints still depend on $v_j$. Second, this velocity-dependency can be eliminated such that 
\begin{align}
&  \frac{D_{s,j}}{v_\text{ref}} \dot{\delta}^\text{min}  \leq \delta_{j+1} - \delta_j \leq \frac{D_{s,j}}{v_\text{ref}} \dot{\delta}^\text{max},~j=0,\dots,N-2,\label{eq_deltad_rateCstrts_III}
\end{align}
where, $v_\text{ref}>0$ is introduced as a reference velocity parameter choice. Note that the approximation in Eqn.  \eqref{eq_deltad_rateCstrts_III} is the exact equivalent of Eqn. \eqref{eq_deltad_rateCstrts} for the zero-reference error setting $e_{y,j}=0$, $e_{\psi,j}=0$, $v_j=v_\text{ref}$. The formulation in Eqn. \eqref{eq_deltad_rateCstrts_III} bears the advantage of eliminating state dependency and is therefore the preferred form for spatial rate constraints. In practice, bounds are employed that are constant over the spatial planning horizon, but in general time-varying in any closed-loop receding horizon control-setting and dependent on the vehicle's operating point. 

To summarise this section, the transformation of time-dependent control rate constraints in Eqn. \eqref{eq_time_rateCstrts} to the path-aligned coordinate frame is not trivial and to be considered as the main disadvantage of a spatial-based system representation. The solution proposed here has opted for the simple state-independent form in Eqn. \eqref{eq_deltad_rateCstrts_III} for linearly constrained optimisation problems, and discussed the role of $v_\text{ref}>0$ as a transformation parameter. 

Finally, an important analogy needs to be made between clothoid and continuous curvature path planning methods to further justify the above approach for the handling of rate constraints in the spatial domain. Continuous curvature and clothoid based path planning methods, such as in \cite{sabelhaus2013using}, typically assume bounds on 'curvature' and 'sharpness' (the change in curvature) resulting in piecewise-affine curvature profiles along the planned path. Note that this implicitly also assumes an approximation of steering rate constraints.

\subsection{Main Two-Step Hierarchical Algorithm}

The following high-level algorithm is proposed to address both Problem \ref{problem1} and \ref{problem2}:

\begin{itemize}
\item[] Pre-processing: Determine an edgy path segment.
\item[1.] Fit an approximating path that shall serve as reference in the subsequent step.
\item[2.] Solve a linear program (LP) based on spatial transformation of nonholonomic vehicle dynamics and the reference path from Step 1 to generate a smoothed path.
\item[] Post-processing: Replace the edgy path segment with the smoothed path.
\end{itemize}
  
An edgy path segment can either be a headland path segment or a headland-to-mainfield lane transition. In all of the following the term \emph{edgy path segment} shall refer to path segments with sharp corners or abrupt directional changes. Edgy path segments are identified by heuristics, in particular, by searching for abrupt directional changes above a certain threshold as well as the labeling of path coordinates as headland path or mainfield lane and detecting transitions between the two with sharp directional changes. 

\begin{remark}\label{def_remark1}
The focus here is on linear programming. In general, also any other objective function can be possible, for example, a quadratic program. However, as shown below a linear programming is sufficient and permits to formulate a hyperparameter-free objective function.
\end{remark}

The next two subsections discuss differences of Step 1 and Step 2 for the two Problems \ref{problem1} and \ref{problem2}.

\subsection{Problem 1: Headland path smoothing\label{subsec_probl1}}

\begin{figure*}
\centering
\begin{subfigure}[t]{.5\textwidth}
  \centering
  \includegraphics[width=.99\linewidth]{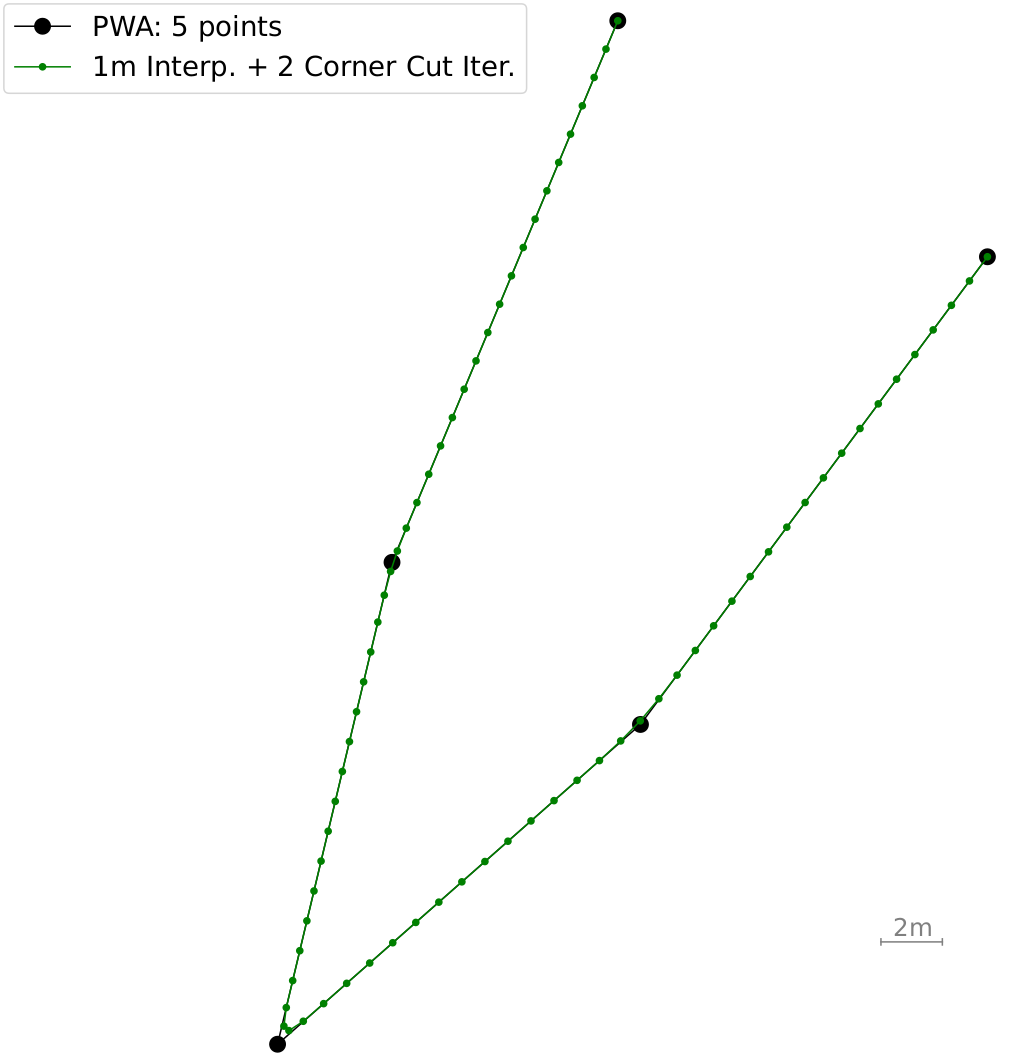}
\caption{Step 1: reference path generation}
  \label{fig_prob1_2steps_a}
\end{subfigure}%
\begin{subfigure}[t]{.5\textwidth}
  \centering
  \includegraphics[width=.99\linewidth]{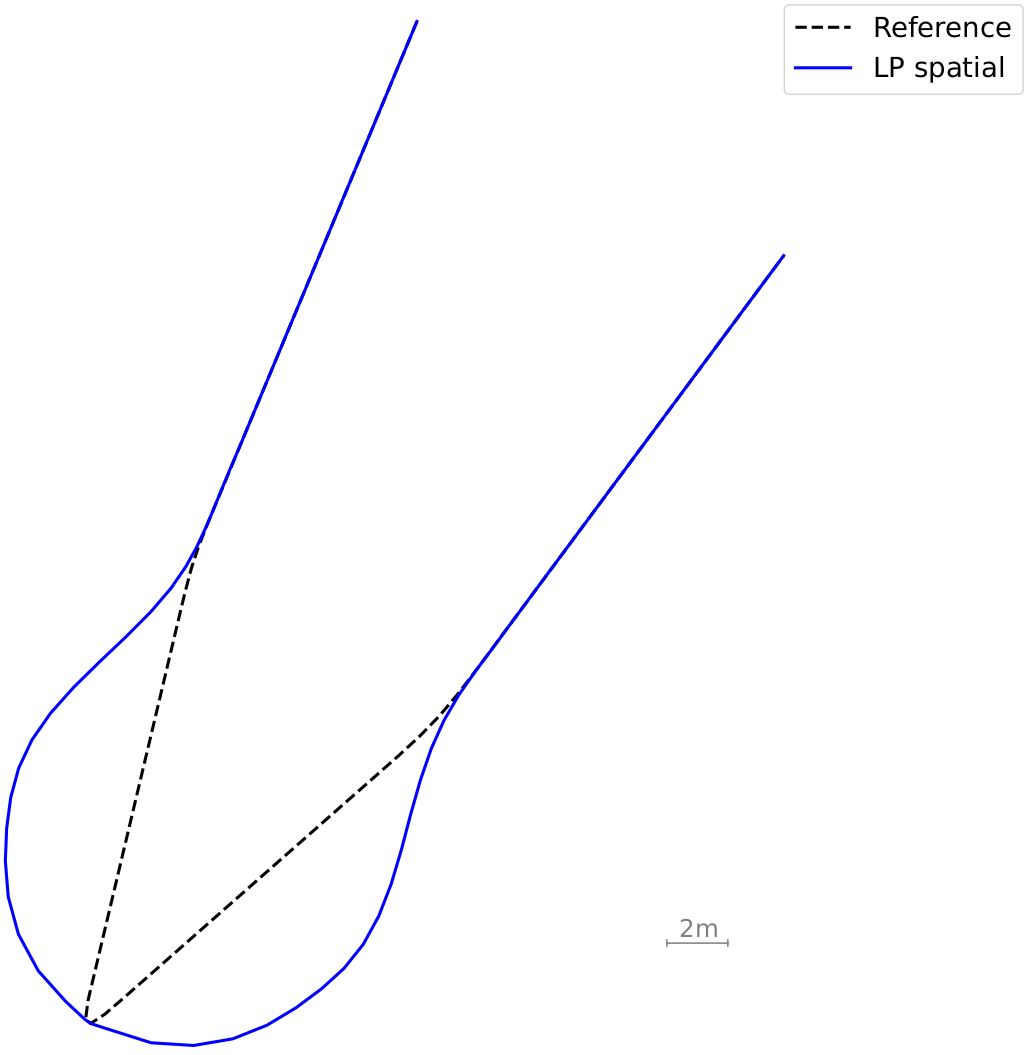}
\caption{Step 2: linear programming solution}
  \label{fig_prob1_2steps_b}
\end{subfigure}
\caption{Problem \ref{problem1}: headland path smoothing. Illustration of two hierarchical steps: (a) generation of a reference (a 5-point piecewise-affine fit that is subsequently interpolated and smoothed via two corner cut-iterations), before (b) using that reference in combination with a spatial domain transformation for a linear program.}
\label{fig_prob1_2steps}
\end{figure*}

In Fig. \ref{fig_f0} an illustrative edgy headland path segment is labeled with \emph{Problem Type 1}. The two steps of the hierarchical algorithm for Problem \ref{problem1} are visualised in Fig. \ref{fig_prob1_2steps}: (a) reference path generation and (b) linear programming solution. 

In a first step, the segment is scaled by heuristics and replaced by a 5-point piecewise-affine (PWA) path as shown in Fig. \ref{fig_prob1_2steps_a} under certain conditions. First, the key characteristic of this scaling is that the tip of the resulting 5-point PWA path shall ensure zero spraying gap. Thus, the distance between the tip and its projection point along the field contour must be less or equal than half the machinery operating width $w$. Secondly, beyond this key characteristic, the shaping of the reference path is a heuristic and free choice. Here, a 5-point PWA path segment is used because it \emph{rounds out} the original edgy headland path segment (more than, for example, an alternative 3-point PWA path segment would) and thereby helps the downstream linear program formulation by providing an improved reference around which the system dynamics are linerised and discretised. Thirdly, the heuristic nature of this first reference path generation step is acknowledged and a potential avenue of future research. Lastly, and as a detail, the popular Chaikin's algorithm (see \cite{chaikin1974algorithm}) is not used for corner cutting and further smoothing of the reference path, since this would replace every edge point with two edge points at each iteration step. Furthermore, it would cut off a significant portion of the tip of the 5-point PWA line after the first iteration. 
  
In a second step, using the 5-point PWA path from Step 1 as a  reference path, the following basic LP is proposed: 
\begin{subequations}
\label{eq_LP1}
\begin{align}
\min\limits_{ \{ \delta_j \}_{j=0}^{N-1}} &\ \  \sum\nolimits_{j=1}^N |e_{y,j}- e_{y,j}^\text{ref} | \label{eq_LP1_objFcn}\\
\mathrm{s.t.} &\  \ e_{y,j} - e_{y,j}^\text{ref} \leq 0, \ j = 1,\dots,N, \label{eq_LP1_eymax_cstrts}\\
&\ \ D_{s,j} \frac{\dot{\delta}_\text{min}}{v_\text{ref}}  \leq \delta_{j+1} - \delta_j \leq D_{s,j} \frac{\dot{\delta}_\text{max}}{v_\text{ref}},  \ j = 0,\dots,N-2,\label{eq_LP1_ddelta_cstrts}\\ 
&\ \ \delta_\text{min} \leq  \delta_j \leq \delta_\text{max},  \ j = 0,\dots,N-1,\label{eq_LP1_delta_cstrts}
\end{align}
\end{subequations}
with $D_{s,j}=s_{j+1}-s_j,\forall j = 0,\dots,N-2$.

Steering actuation constraints are in Eqn. \eqref{eq_LP1_ddelta_cstrts} and \eqref{eq_LP1_delta_cstrts} as discussed in Sect. \ref{subsec_spatial}. In practice, constraints in Eqn. \eqref{eq_LP1_eymax_cstrts}, which enforces the resulting path to stay on one side of the reference path, is relaxed to, $e_{y,j} - e_{y,j}^\text{ref} \leq \sigma $, by introducing a non-negative slack variable $\sigma\geq 0$ and adding it to the objective function \ref{eq_LP1_objFcn} with a high weight such that $\sum\nolimits_{j=1}^N |e_{y,j}- e_{y,j}^\text{ref} |$ is replaced with $\sum\nolimits_{j=1}^N |e_{y,j}- e_{y,j}^\text{ref} | + 10^{16}\sigma$. Furthermore, for the absolute value arguments, $N$ non-negative surrogate variables are introduced and the set of inequality constraints is extended accordingly. In the end, there are $n_u=2N+1$ scalar real-valued optimisation variables. Typical sizes of the LP (i.e., the number of variables and inequality constraints) are stated in the numerical example Section \ref{sec_IllustrativeEx}.
 
For generality, reference $e_{y,j}^\text{ref}$ is included in Eqn. \eqref{eq_LP1_objFcn} and \eqref{eq_LP1_eymax_cstrts}. In practice, it is typically set to zero, $e_{y,j}^\text{ref}=0,\forall j = 1,\dots,N$. 

One benefit of proposed formulation in LP \eqref{eq_LP1} is that it is hyperparameter-free. There are no weighting parameters.

An even smaller alternative LP was considered, with a minmax-objective function, $\min\nolimits_{ \{ \delta_j \}_{j=0}^{N-1}} \  \max\nolimits_{ j\in\{1,\dots,N\}} |e_{y,j}- e_{y,j}^\text{ref} |$, replacing Eqn. \eqref{eq_LP1_objFcn}. However, this was found to be unsuited since only the maximum deviation from the reference path would be penalised, which resulted in jagged trajectories not accurately tracking the complete reference path.

After the solution of LP \eqref{eq_LP1}, a smoothed headland path segment as shown in Fig. \ref{fig_prob1_2steps_b} results.
 
Finally, \emph{LP-iterations} or refinement steps can be conducted. Thus, for the first solution of LP \eqref{eq_LP1} a 5-point-PWA path is used as a reference. Then, the resulting path is used as a reference for the second solution of LP \eqref{eq_LP1}. In theory, many such LP-iterations can be carried out. In practice, one refinement step was found to be useful and sufficient.

\subsection{Problem 2: Headland-to-mainfield lanes transitions}

Before stating the LP for Problem \ref{problem2}, the term \emph{headland-to-mainfield lanes} transitions is clarified. It is used comprehensively for brevity but shall always refer to both, (i) transitions from a headland to a mainfield lane, but also (ii) transitions from a mainfield lane to the headland path. Both types of transitions naturally occur for a field coverage path. As will be shown below, a subtle difference in constraints formulation is applied to differentiate the two types.

The following basic LP is proposed for Problem \ref{problem2}: 
\begin{subequations}
\label{eq_LP2}
\begin{align}
\min\limits_{ \{ \delta_j \}_{j=0}^{N-1}} &\ \  \sum\nolimits_{j=1}^N |e_{y,j}- e_{y,j}^\text{ref} | \label{eq_LP2_objFcn}\\
\mathrm{s.t.} &\ \ D_{s,j} \frac{\dot{\delta}_\text{min}}{v_\text{ref}}  \leq \delta_{j+1} - \delta_j \leq D_{s,j} \frac{\dot{\delta}_\text{max}}{v_\text{ref}},  \ j = 0,\dots,N-2,\\ 
&\ \ \delta_\text{min} \leq  \delta_j \leq \delta_\text{max},  \ j = 0,\dots,N-1,
\end{align}
\end{subequations}
with $D_{s,j}=s_{j+1}-s_j,\forall j = 0,\dots,N-2$.

In practice, the objective function Eqn. \eqref{eq_LP2_objFcn} is slightly refined to better avoid undesired overshoots out of the mainfield area and into the headland area. Therefore, Eqn.  \eqref{eq_LP2_objFcn} is replaced with $\sum\nolimits_{j=1}^N c_j |e_{y,j}- e_{y,j}^\text{ref} |$ and weights   
\begin{equation*}
c_j=\begin{cases} 100 & j=1,\dots,I \\ 1 & j=I+1,\dots,N
\end{cases}
\end{equation*}
for transitions from headland to a mainfield lane, and
\begin{equation*}
c_j=\begin{cases} 1 & j=1,\dots,I-1 \\ 100 & j=I,\dots,N
\end{cases}
\end{equation*}
for transitions from mainfield lane to the headland path, where index $I$ is computed from the reference path as its last and first index within a 1 m radius to the headland path for the two transition types. The value 100 is selected exemplarily and can be any large constant, such that the objective function of the LP can still be considered hyperparameter-free. The benefit of this weighted approach is that no additional inequality constraints need to be introduced. As a result, LP \eqref{eq_LP2} for Problem \ref{problem2} is smaller than LP \eqref{eq_LP1} for Problem \ref{problem1}. 

As for above LP \eqref{eq_LP1}, for the absolute value arguments in Eqn. \eqref{eq_LP2_objFcn}, $N$ non-negative surrogate variables are introduced and the set of inequality constraifnts is extended accordingly. In the end, there are $n_u=2N$ scalar real-valued optimisation variables. Typical sizes of the LP (i.e., the number of variables and inequality constraints) are stated in the numerical example in Section \ref{sec_IllustrativeEx}. As in the previous subsection, reference $e_{y,j}^\text{ref}$ in LP \eqref{eq_LP2} is typically set to zero, $e_{y,j}^\text{ref}=0,\forall j = 1,\dots,N$.

\begin{figure*}
\centering
\begin{subfigure}[t]{.33\textwidth}
  \centering
  \includegraphics[width=.99\linewidth]{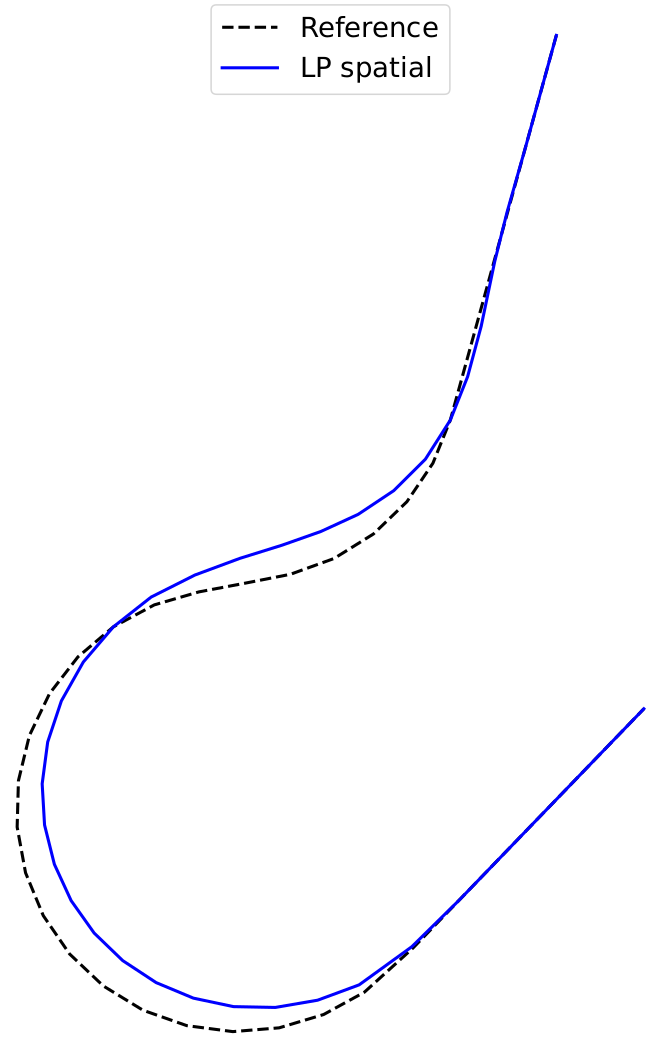}
\caption{$R_\text{Dubins}=5$m}
  \label{fig_prob2_3R_sub1}
\end{subfigure}%
\begin{subfigure}[t]{.33\textwidth}
  \centering
  \includegraphics[width=.99\linewidth]{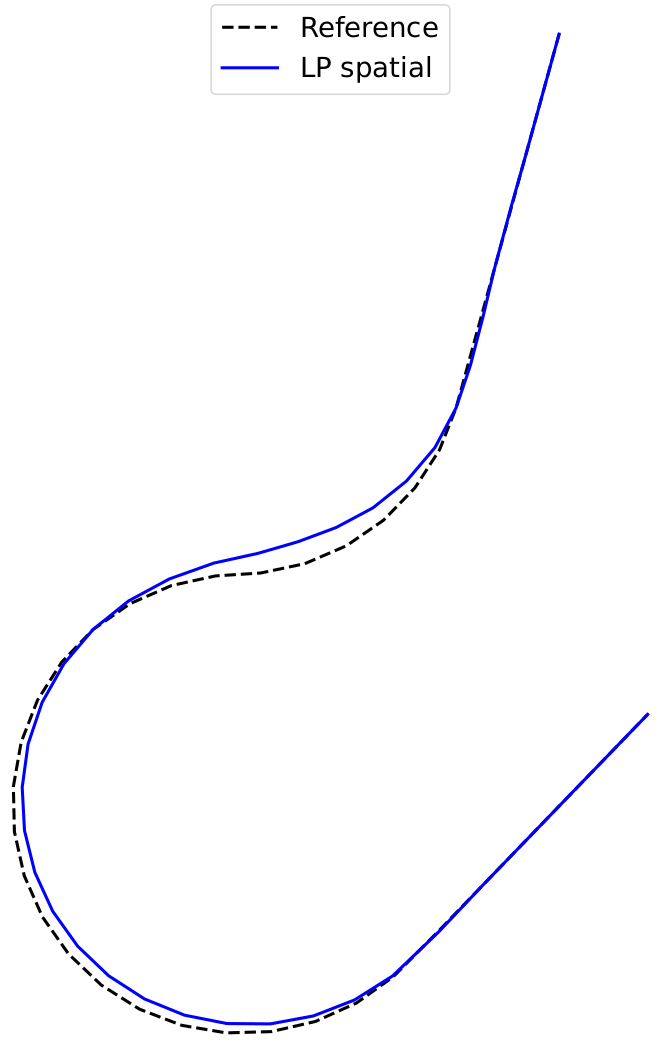}
\caption{$R_\text{Dubins}=5.33$m}
  \label{fig_prob2_3R_sub2}
\end{subfigure}
\begin{subfigure}[t]{.33\textwidth}
  \centering
  \includegraphics[width=.99\linewidth]{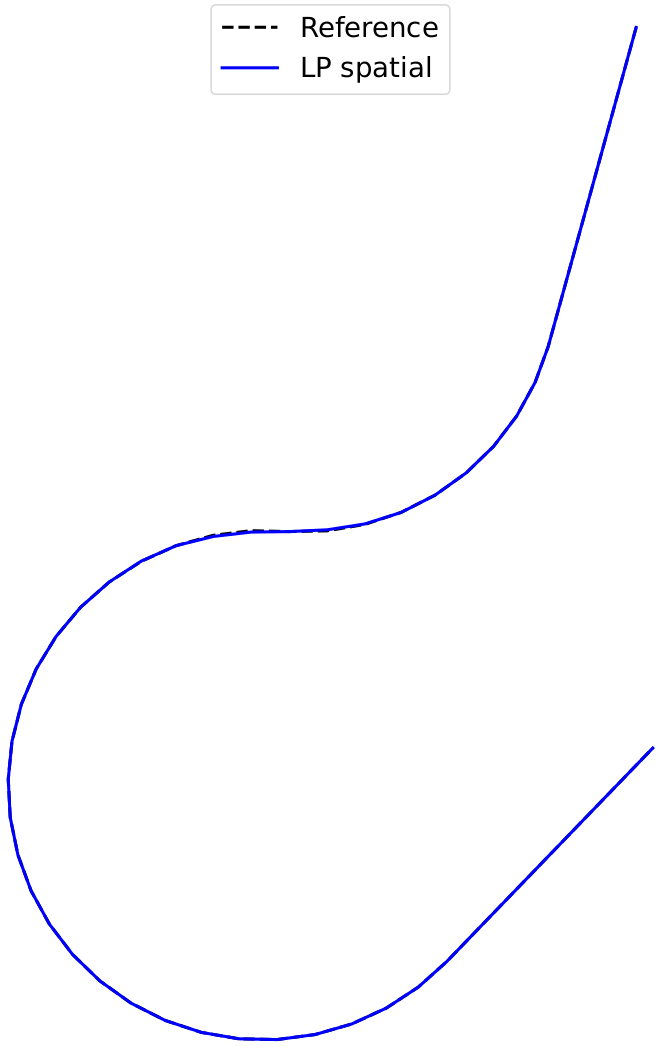}
\caption{$R_\text{Dubins}=7$m}
  \label{fig_prob2_3R_sub3}
\end{subfigure}
\caption{Problem \ref{problem2}: The effect on tracking performance when using three different $R_\text{Dubins}$ in the reference path generation step is visualised. Paths denoted with 'Reference' and 'LP spatial' denote the paths before and after the LP-solutions, respectively.}
\label{fig_prob2_3R}
\end{figure*}

Figure \ref{fig_prob2_3R} visualises the result of a headland-to-mainfield lane transition. As part of Step 1 of the hierarchical algorithm for Problem \ref{problem2}, a Dubins path is fitted and replaces the edgy original transition. The Dubins path, plus an additional small path segment before and after the Dubins path to permit additional maneuvering space, serves as reference to Step 2 of the hierarchical algorithm based on which LP \eqref{eq_LP2} is formulated and solved.

\begin{figure*}
\centering
\begin{subfigure}[t]{.5\textwidth}
  \centering
  \includegraphics[width=.99\linewidth]{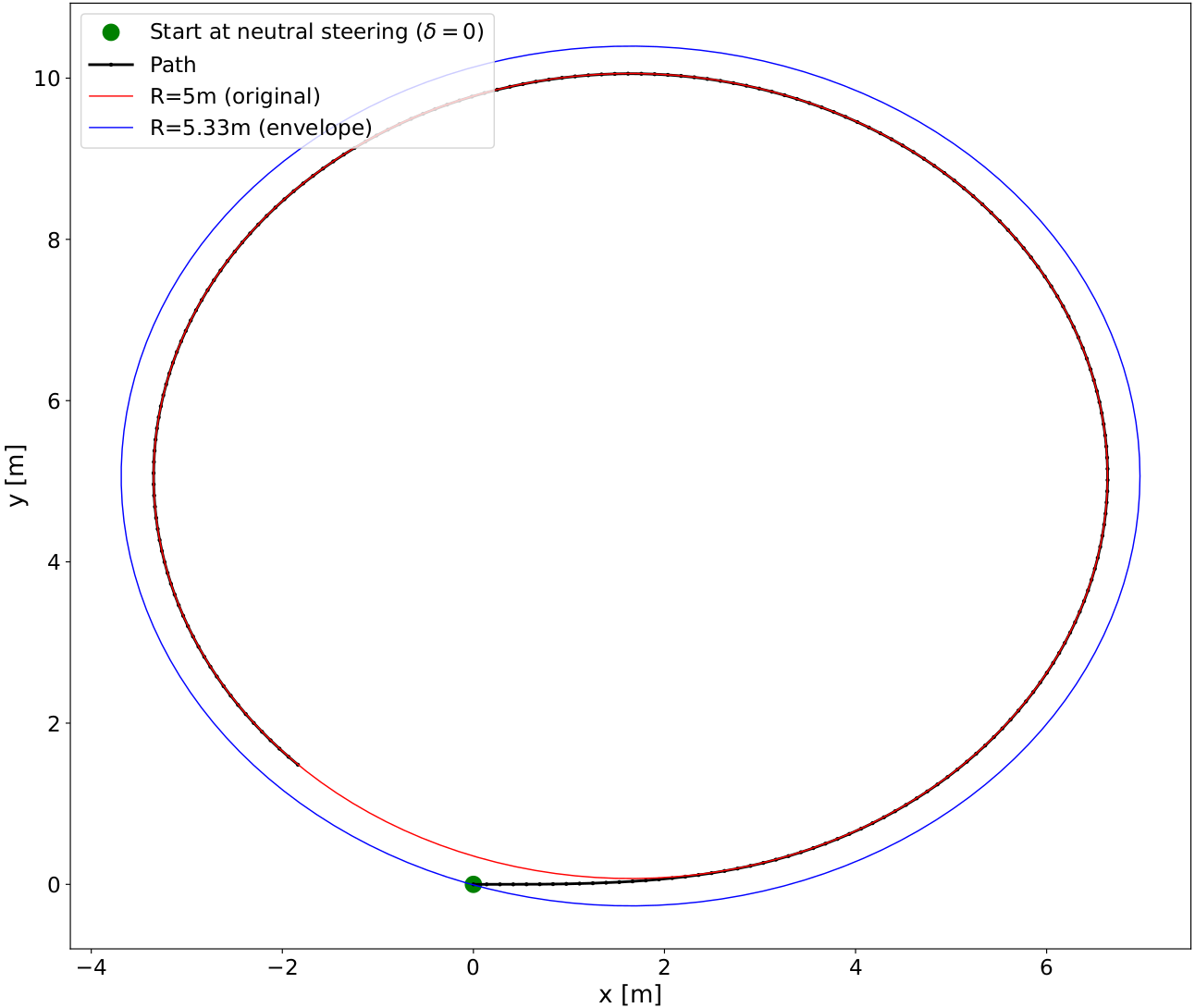}
\caption{Simulation 1}
  \label{fig:sub1}
\end{subfigure}%
\begin{subfigure}[t]{.5\textwidth}
  \centering
  \includegraphics[width=.99\linewidth]{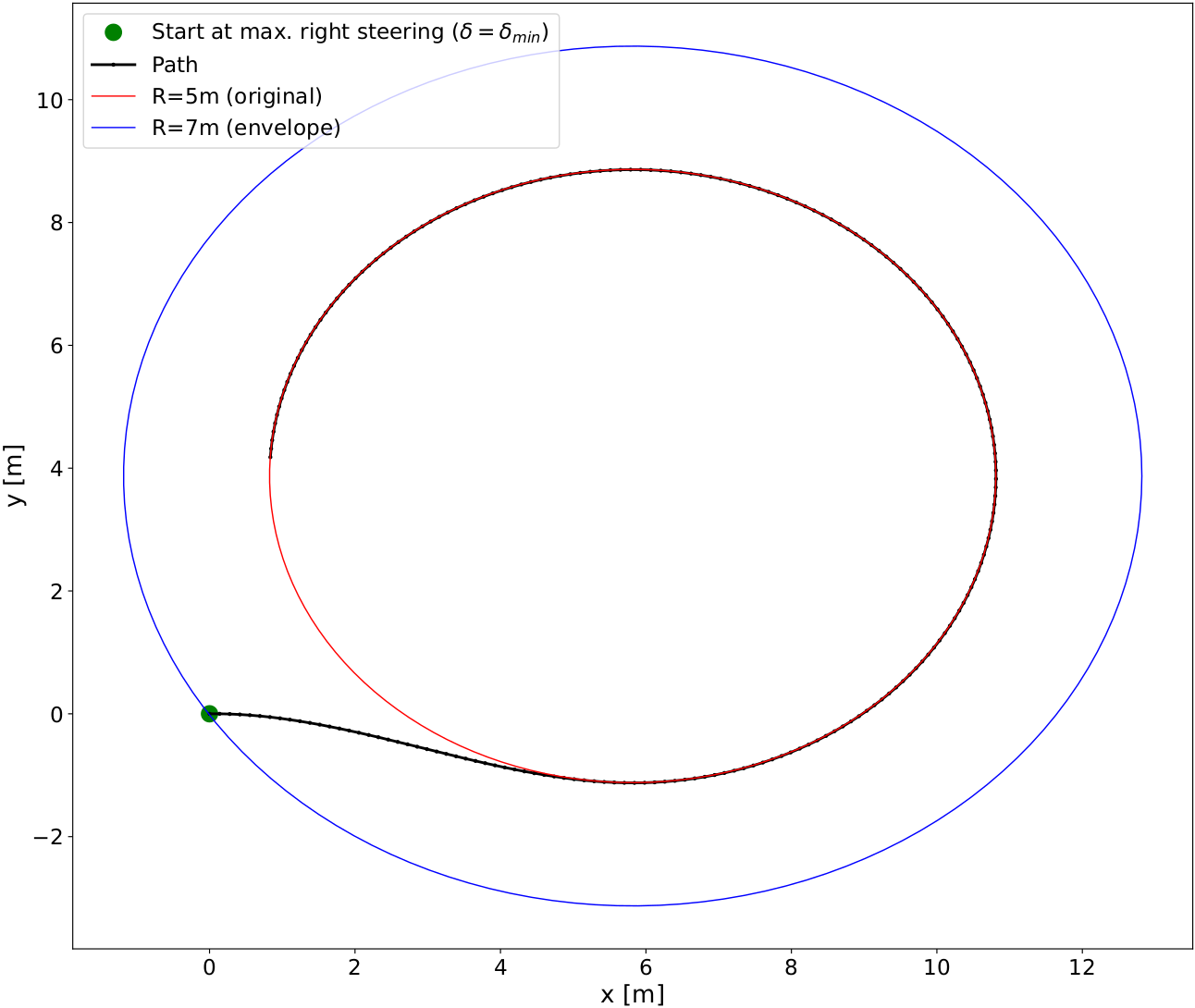}
\caption{Simulation 2}
  \label{fig:sub2}
\end{subfigure}
\caption{Given an initial vehicle position $[x(t),y(t),\psi(t)]=[0,0,0]$ and either $\delta(t)=0$ or $\delta(t)=\delta_\text{min}$ as initial steering at time $t=0$, two simulations are conducted, applying maximum possible steering at each sampling time according to the law, $\delta(t+T_s)=\max\{\delta(t)+T_s \dot{\delta}_\text{max},\delta_\text{max}\}$ for all $t>0$. The effect of steering rate constraints on the transition phase, before the vehicle reaches its minimum turning radius, is visualised. Enveloping circles are fitted. The effect of using different radii of the envelope circles for reference path generation is visualised in Fig. \ref{fig_prob2_3R}.}
\label{fig_prob2_2sim}
\end{figure*}

Figures \ref{fig_prob2_2sim} and \ref{fig_prob2_3R} are of special interest for two reasons. First, they illustrate how Dubins paths are not suited for the generation of precision paths to be followed by autonomous machines online, especially when calculating Dubins paths with minimal turning radius of the machine (see Fig. \ref{fig_prob2_3R_sub1}). Due to actuation constraints, deviations would be expected in any online tracking application. Second, it illustrates the effect on tracking performance when using different turning radii $R_\text{Dubins}$ in the Dubins reference path generation step. When using the minimal turning radius, results in the deviation shown in Fig. \ref{fig_prob2_3R_sub1} result. By using an artificially larger $R_\text{Dubins}$, the LP-solution trajectory more closely follows the reference path, see Fig. \ref{fig_prob2_3R_sub2} and \ref{fig_prob2_3R_sub3}. The drawback of a larger $R_\text{Dubins}$ is that the resulting path is longer and a larger area of compaction will occur from corresponding tyre traces and the corresponding steering maneuvers, which reduces the productive crop area in the field. A trade-off with a preference towards small $R_\text{Dubins}$ for the reference path generation step is recommended.
 
Finally, in contrast to Problem \ref{problem1} and its discussion in Section \ref{subsec_probl1} for the case of Problem \ref{problem2}, no LP-iterations are conducted. It was empirically found that they did not improve results and are omitted.

\section{Methods}

\begin{figure}
\centering
\includegraphics[width=8.5cm]{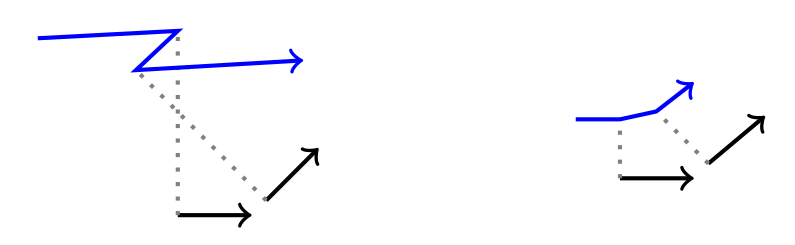}
\caption{Sketch illustrating the danger of jaggedness for spatial-based optimisation. A reference path (black) is desired such that lateral deviations resulting from the LP-solution (blue) remain small. This is the case in the right drawing, but not in the left drawing.}
\label{fig_f11}
\end{figure}

This section elaborates on important parameter choices and other practical details. First, the selection of discretisation spacing, $D_{s,j}=s_{j+1}-s_j,~\forall j=0,\dots,N-1$, is discussed, which typically is selected as uniformly spaced but can also be non-uniformly spaced. Its importance for spatial-based numerical optimisation is underlined. Its selection influences both the size of the LPs (the larger the spacing the smaller the LP) and the feasibility and quality of resulting trajectories. The smaller the spacing, the more danger there is of generating jaggedness as sketched in Fig. \ref{fig_f11}, whereas the larger the spacing the more the  resulting trajectories obtain piecewise-affine and thus an edgy shape. The danger of generating jaggedness, as sketched in Fig. \ref{fig_f11}, is a disadvantage of the spatial-based optimisation problem formulation. On the other hand, in the case of typical uniform spacing for a constant $D_s>0$ and $D_{s,j}=D_s,~\forall j=0,\dots,N-1$, there is only a single hyperparameter that can be well tuned. In practice, this can be done either by trial-and-error for a given field or by a worst-case simulation analysis for given system dynamics and actuation limits. In case of in-field path planning, where there typically are no fast-moving obstacle avoidance requirements and precision path trajectories are calculated offline, worst-case simulation analysis reduces to looping over complex Dubins reference paths, calculating spatial-based LP-tracking solutions and evaluating feasibility and tracking error for a given discretisation spacing candidate $D_s>0$. For the numerical experiments in the next section $D_s=1$ m is set for all instances of both Problem \ref{problem1} and \ref{problem2}.

Second, for Problem \ref{problem1} a comment about the generation of a reference path  is made. The presented 5-point PWA heuristic with tip-scaling serves the objective of spraying gap avoidance. However, it may not be optimal with respect to pathlength minimisation. For example, a simpler 3-point PWA heuristic might also be used for a reference path. Heuristic variations on the reference path generation step taking local field contour shape information into account will be the subject of future research.

All experiments were run on a laptop running Ubuntu 22.04 equipped with an Intel Core i9 CPU @5.50GHz×32 and 32 GB of memory. For the solution of the LPs Scipy's (cf. \cite{virtanen2020scipy}) \texttt{linprog}()-solver in default settings was used.

\begin{table}
\vspace{0.3cm}
\caption{Parameters used throughout all numerical experiments. The value $\delta_\text{max}=31$deg was set such that for a minimum turning radius of $R_\text{Dubins}=5$m, it holds $R_\text{Dubins}=\frac{l}{\text{tan}(\delta_\text{max})}$.}
\centering
\begin{tabular}{l|llll}
\hline \hline
Parameter & $l$ & $\delta_\text{max}$ & $\dot{\delta}_\text{max}$ & $v_\text{ref}$ \\[1pt]
\hline
Value & 3m & 31deg  & 15deg/s  & 5km/h \\
\hline
\end{tabular}
\label{tab_param}
\end{table}

\section{Results\label{sec_IllustrativeEx}}

\begin{figure}
\centering
\includegraphics[width=8.5cm]{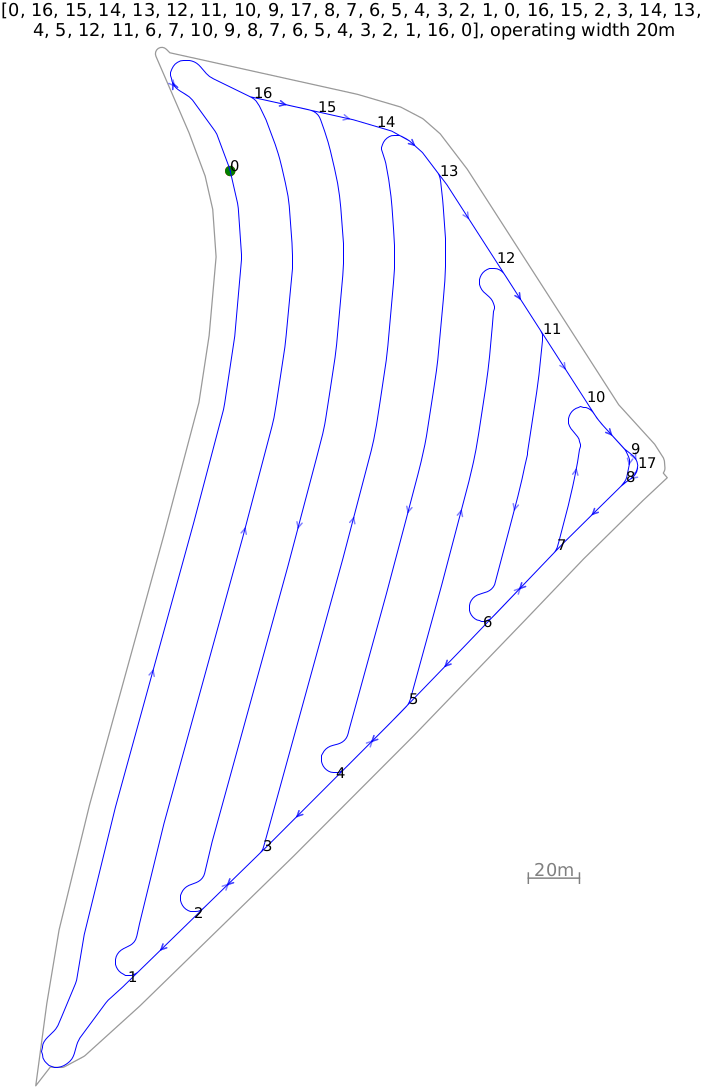}
\caption{Experiment: 3 instances of Problem \ref{problem1} and 16 instances of Problem \ref{problem2} needed to be solved. The path obtained from proposed method is visualised. The sequence of nodes, $[0,16,\dots]$, gives rise to a full field coverage path plan and is calculated by the routing algorithm from \cite{plessen2019optimal} subject to the constraint that a full initial headland coverage (from node $0$ to node $0$ along the headland path before servicing the mainfield lanes) is sought as part of the full area coverage path. The field entrance is denoted by node $0$. The inter-lane distance is 20 m. Tracing the number sequence in the heading gives rise to the area coverage path.}
\label{fig_f8}
\end{figure}

\begin{figure}
\centering
\includegraphics[width=8.5cm]{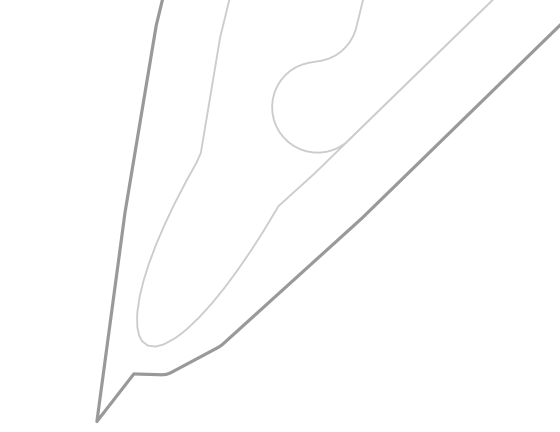}
\caption{Illustration of a zoom-in for an alternative solution approach to Problem \ref{problem1}: a Bezier curve of third order is fitted to the 5-point PWA path. This approach does not explicity account for vehicle actuation constraints. For comparison, see the bottom turn in Fig. \ref{fig_f8} and \ref{fig_f10}, which results from the solution of LP \eqref{eq_LP1} and takes vehicle actuation constraints into account.}
\label{fig_f9}
\end{figure}

\begin{figure}
\centering
\includegraphics[width=8.5cm]{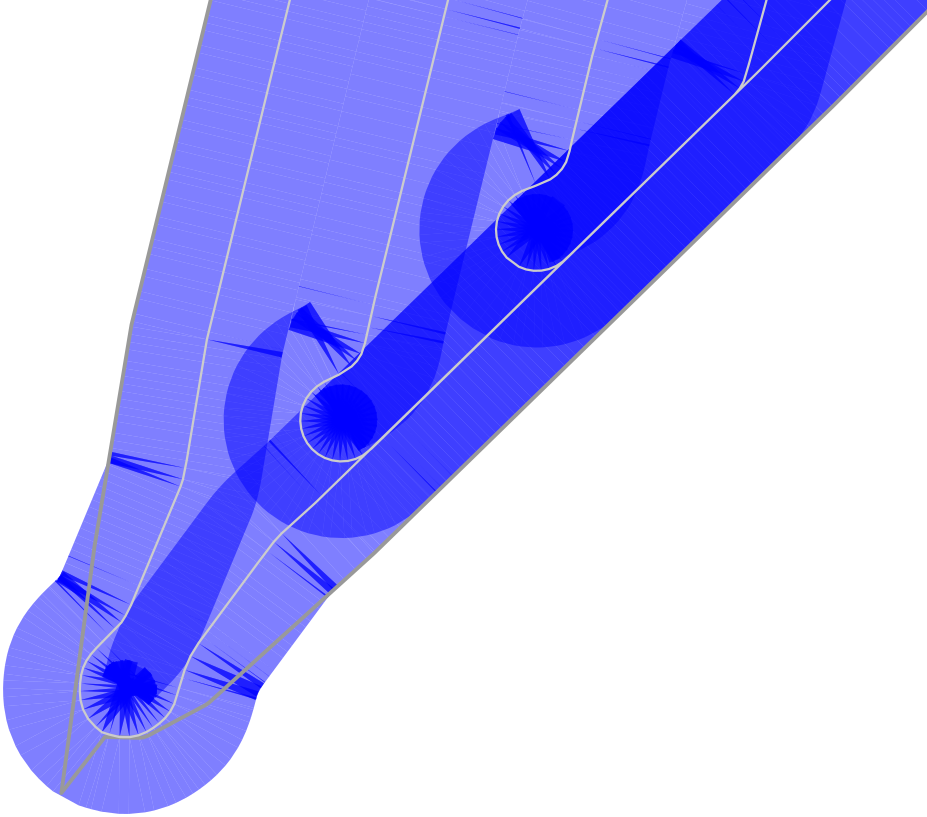}
\caption{Zoom-in displaying sprayed coverage area (blue) for the path plan of Fig. \ref{fig_f8}. Darker blue areas imply overlapping spray area. In this illustrating simulation no section control with individual switching of spray nozzles is assumed (which could in practice reduce the overlapping area).}
\label{fig_f10}
\end{figure}

\begin{table}
\vspace{0.3cm}
\caption{Average number of constraints $\bar{n}_\text{cstrts}$ and average number of variables $\bar{n}_\text{u}$ of the LPs, and corresponding average solution runtimes for the LP- and Bezier-approach for 3 instances of Problem \ref{problem1} in Figure \ref{fig_f8}.}
\centering
\begin{tabular}{ccc|c}
\hline \hline
\rowcolor[gray]{1.0} $\bar{n}_\text{cstrts}$ & $\bar{n}_\text{u}$ & $\bar{\tau}_\text{LP}$ & $\bar{\tau}_\text{Bezier}$ \\[1pt] 
\hline
294 & 149 & 0.008 s & 0.002 s \\
\hline
\end{tabular}
\label{tab_prob1}
\end{table}

\begin{table}
\vspace{0.3cm}
\caption{Average number of constraints $\bar{n}_\text{cstrts}$ and average number of variables $\bar{n}_\text{u}$ of the LPs, and corresponding average solution runtimes for the LP- and Dubins-solution for 16 instances of Problem \ref{problem2} in Figure \ref{fig_f8}.}
\centering
\begin{tabular}{ccc|c}
\hline \hline
\rowcolor[gray]{1.0} $\bar{n}_\text{cstrts}$ & $\bar{n}_\text{u}$ & $\bar{\tau}_\text{LP}$ & $\bar{\tau}_\text{Dubins}$ \\[1pt] 
\hline
98 & 50 & 0.0017 s & 0.0004 s \\
\hline
\end{tabular}
\label{tab_prob2}
\end{table}

\begin{table}
\caption{Average maximum absolute deviation from a Dubins reference path, $\text{mean} \left\{e_\text{y,abs}^\text{max} \right\}$, and maximum maximum absolute deviation, $\max \left\{e_\text{y,abs}^\text{max} \right\}$, over 16 instances of Problem \ref{problem2} for three different $R_\text{Dubins}\in\{5\text{m},5.33\text{m},7\text{m}\}$. See Fig. \ref{fig_prob2_3R} for visualisation of one of such instances.}
\vspace{0.3cm}
\centering
\begin{tabular}{lll}
\hline \hline
\rowcolor[gray]{1.0} $R_\text{Dubins}$ & $\text{mean} \left\{e_\text{y,abs}^\text{max} \right\}$ & $\max \left\{e_\text{y,abs}^\text{max} \right\}$ \\[1pt] 
\hline
5 m & 0.32 m & 0.97 m \\
5.33 m & 0.16 m & 0.48 m \\
7 m & 0.01 m & 0.04 m \\
\hline
\end{tabular}
\label{tab_ey}
\end{table}

For numerical evaluation the field from Fig. \ref{fig_problFormulation} is considered. It has a size of 4.5 ha and requires the solution of 3 instances of Problem \ref{problem1} and 16 instances of Problem \ref{problem2}. The field under consideration involves freeform mainfield lanes, and multiple tight edges for transitions to and from the mainfield lanes and along the headland path.

The solution of proposed method is presented in Fig. \ref{fig_f8}. Parameters used throughout are stated in Table \ref{tab_param}. Fig. \ref{fig_f9} shows a zoom-in for an alternative solution approach to Problem \ref{problem1}. Here, a Bezier-curve (\cite{pastva1998bezier}) of third order is fitted to the 5-point PWA path. This implies solving a least-squares problem to identify parameters before the evaluation of a Bernstein polynomial to generate the Bezier curve-fit. A benefit is the simplicity of this approach. A disadvantage is that even though the interpolating fit is smooth, this method does not take vehicle actuation limits into account. Therefore, the path is in general not feasible to drive. In contrast, proposed LPs \eqref{eq_LP1} and \eqref{eq_LP2} do account for vehicle actuation limits.

The solution times and sizes of LPs for Problem \ref{problem1} and \ref{problem2} are in Tables \ref{tab_prob1} and \ref{tab_prob2}, respectively. Sizes of the LPs are typically larger for  Problem \ref{problem1}, with a larger average number of constraints ($\bar{n}_\text{cstrts}$) and a larger average number of variables ($\bar{n}_\text{u}$). The solution times are very small, which is encouraging for the general approach. 

\begin{figure*}
\begin{center}
\vspace{0.0cm}
\bgroup
\def\arraystretch{1}
\begin{tabular}{cc}
 %
\begin{minipage}{0.5\textwidth}\vspace*{0.1cm}\includegraphics[width=0.99\linewidth]{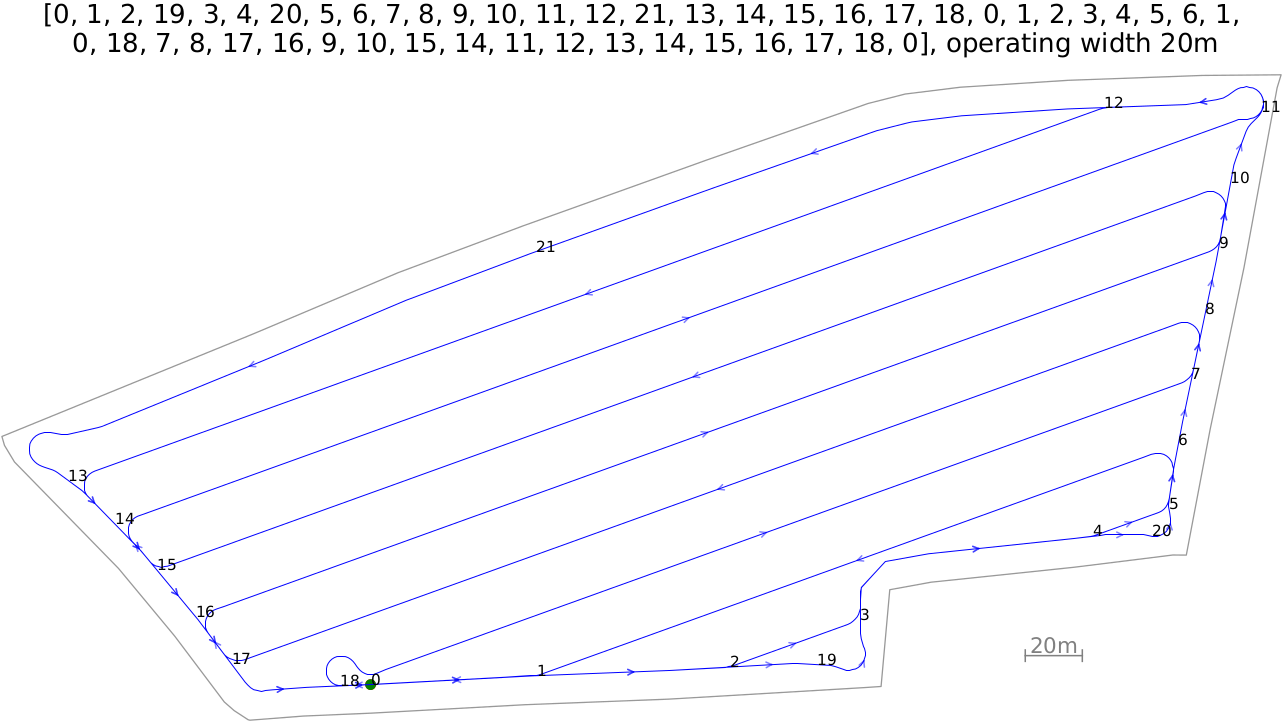}\end{minipage}  & 
\begin{minipage}{0.5\textwidth}\vspace*{0.1cm}\includegraphics[width=0.99\linewidth]{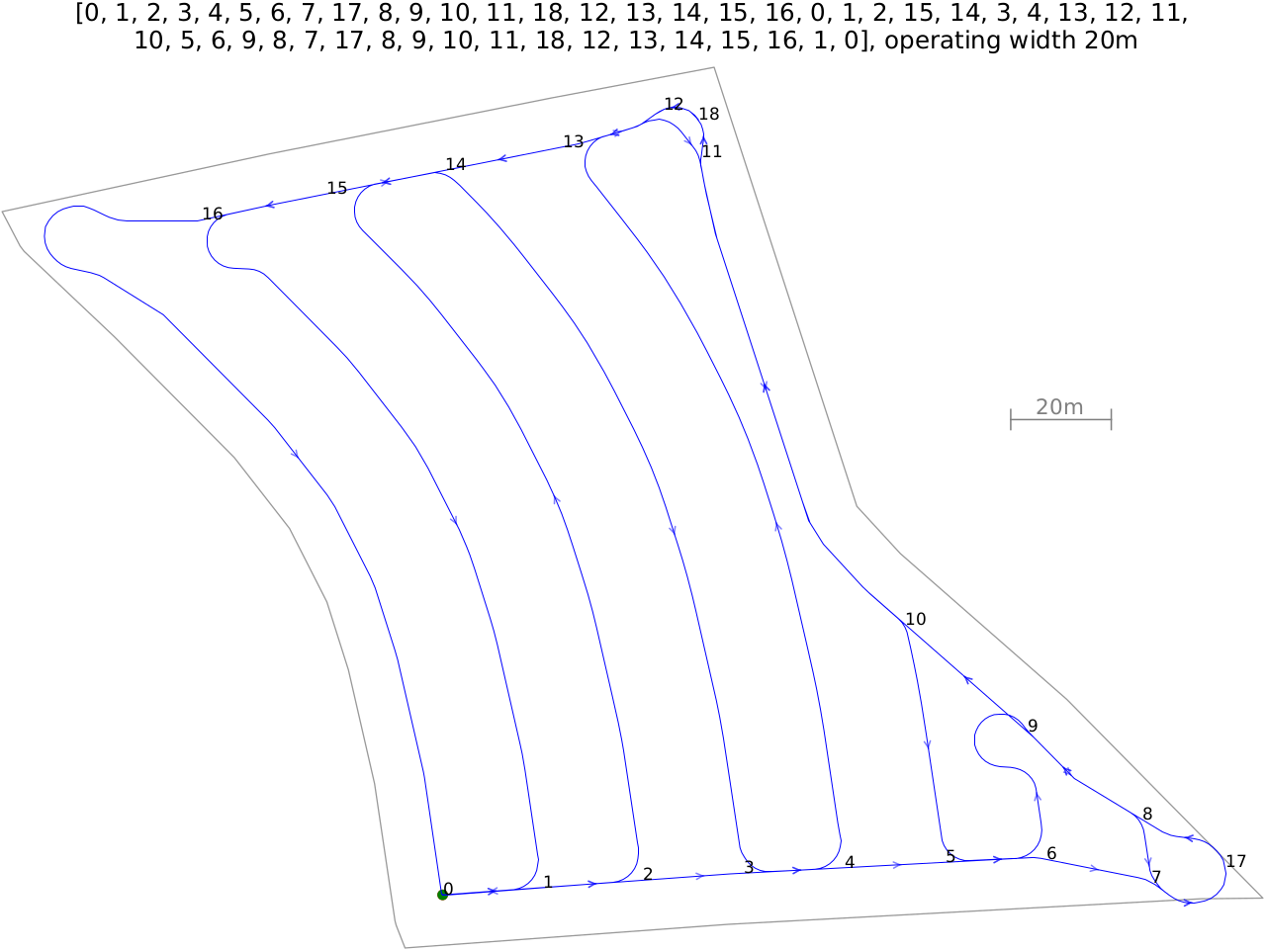}\end{minipage}\\
 %
\begin{minipage}{0.4\textwidth}\vspace*{0.3cm}\includegraphics[width=0.99\linewidth]{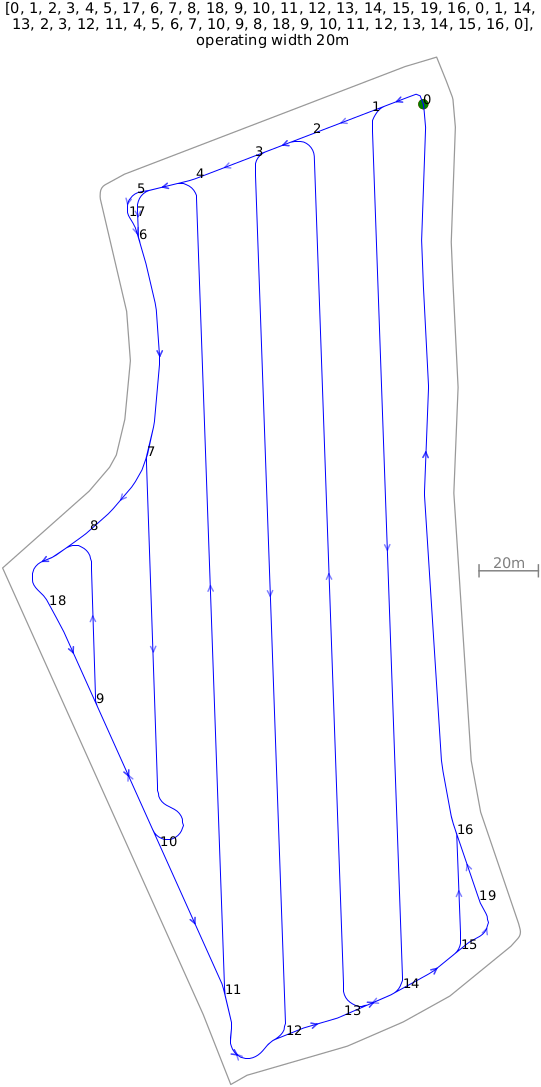}\end{minipage}  & 
\begin{minipage}{0.45\textwidth}\vspace*{0.3cm}\includegraphics[width=0.99\linewidth]{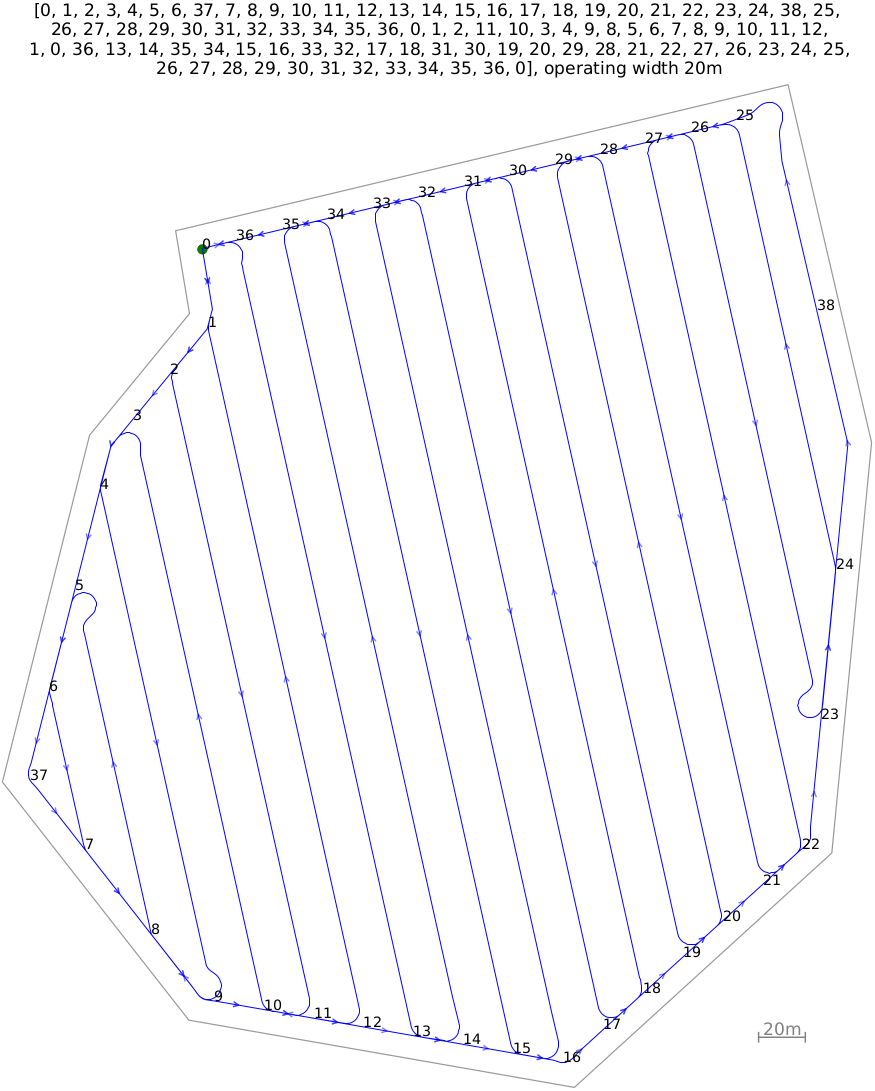}\end{minipage}\\[-10pt]
%
\end{tabular}
\egroup
\end{center}
\caption{More illustrative examples with edgy field contours: 3 fields with straight mainfield lanes and 1 field with freeform curved mainfield lanes. For each example, multiple instances of Problem \ref{problem1} and Problem \ref{problem2} had to be solved. Overall, 16 for Problem \ref{problem1} and 68 for Problem \ref{problem2}.}
\label{fig_4morefields}
\end{figure*}

Figure \ref{fig_f10} illustrates a zoom-in for sprayed coverage area avoiding any area coverage gaps. For three different turning radii, $R_\text{Dubins}\in\{5\text{~m},5.33\text{~m},7\text{~m}\}$, used in the reference path generation step (Step 1 of the hierarchical two-step algorithm), Table \ref{tab_ey} illustrates the effect on lateral deviations between the Dubins-path used for reference path generation and the final smoothed path resulting from the LP-solution when accounting for vehicle actuation limits. The effect of using different $R_\text{Dubins}$ was significant. See also Fig. \ref{fig_prob2_3R} for visualisation. More illustrative examples are provided in Fig. \ref{fig_4morefields}.

\section{Discussion}

In view of precision agriculture and autonomous robot applications an empirical finding of this work was that planning of transitions between headland path and mainfield lanes based only on Dubins-paths was not a viable option. Signification closed-loop tracking errors would be expected. As shown in Fig. \ref{fig_prob2_3R_sub1} there are significant deviations between Dubins path and the path that is actually feasible considering actuation rate limits, especially when using the minimal vehicle turning radius, $R_\text{Dubins}=5\text{~m}$, for Dubins planning.

A general comment about importance of input data quality is made. The quality of results for data-dependent techniques is a function of input data. For very \emph{edgy} field contour input data, as is the case for all 5 fields in Fig. \ref{fig_f8} and \ref{fig_4morefields}, a decision has to be made: (i) achieving full area coverage but exceeding contours when planning a nonholonomic path (see Fig. \ref{fig_f9}), or (ii) permitting area coverage gaps but achieving a path with sprayed area fully inside field contours. Both objectives cannot be achieved simultaneously. According to the formulations of Problem \ref{problem1} and \ref{problem2} avoiding area coverage gaps is the priority of this paper. Objective (ii) is likewise not trivial and a subject for future work.

By counter-examples and discussion, the following provides further supporting arguments for the proposed method.

In \cite{oh2024sharable} a semi-analytic path planner using three consecutive clothoids was proposed. Using continuity conditions and boundary conditions for three clothoids, 8 nonlinear algebraic equations with 15 variables were derived. Using algebraic manipulations these can be reduced to two nonlinear equations for the two unknowns, which can be solved using Newton’s method involving two tuning parameters. This method is interesting because of its analytic approach. However, it has three drawbacks that render it unsuitable for the applications in this paper. The first is that it is based on only three clothoids, which is insufficient to approximate all six combinations of Dubins paths (\cite{dubins1957curves}). To see this by counter-example, consider a RSR-combination (right turning, straight driving, right turning), which requires 4 clothoids for path smoothing. The second disadvantage is that it is not straightforward to correctly set the rate of curvature as a tuning parameter. For example, for Problem Type \ref{problem1}, steering is required that is not always fully operating at the actuating limits. To make a semi-analytic path planner work (after any extension of the approach to all 6 Dubins combinations), scenario heuristics have to be devised and the method thereby evolves into a sampling scheme. The third limitation is that this analytic approach is limited to a specific vehicle model, namely one with no slip conditions at the front and the rear wheels and the centre of gravity being located at the centre of the rear axle.

In contrast, when extending the kinematic bicycle model (\cite{rajamani2011vehicle}) with slip to
\begin{equation}
\begin{bmatrix} \dot{x},~\dot{y},~\dot{\psi},~\dot{v} \end{bmatrix} = \begin{bmatrix} v \cos(\psi + \beta),~ v \sin(\psi + \beta),~ \frac{v \cos(\beta)}{l} \tan(\delta),~ a \end{bmatrix},\label{eq_kinmodel_withslip}
\end{equation}
where $\beta=\text{arctan}\left( \frac{\delta}{2} \right)$ denotes the slip angle, $a$ denotes acceleration control, and the centre of gravity is now located in the middle at distance $\frac{l}{2}$ to both front and rear wheels, then the same conceptual approach described in this paper can be applied. The only difference is the replacement of Eqn. \eqref{eq_dotxypsi_nonlinkinbicmdl} with \eqref{eq_kinmodel_withslip}. See also \cite{miao2022research} for an alternative vehicle model. Similarly, trailer dynamics can be incorporated (equations in \cite{oksanen2004optimal}, \cite{plessen2017reference}) into the proposed methodology. 

Similarly to \cite{oh2024sharable}, \cite{sabelhaus2013using} applied an analytic approach in earlier work in an agricultural context. They presented analytic formulas and visualisations for 9 different symmetric lane-to-lane transitions (\verb!Omega-turn!, \verb!U-turn!, \verb!Gap-turn!, \verb!Slope-turn!, \verb!Fishtail-turn!, \verb!Fishtail-turn with slope!, \verb!Minimal longitudinal width!, \verb!Pinhole-turn!, \verb!Reversal pinhole turn!) for continuous-curvature path planning.  They also presented three simulation experiments with longitudinal, lateral and angle offset of start and end point. Characteristically, all paths were constructed based on path segments with maximum curvature rate (clothoids with so called maximal sharpness) and circle segments with maximum curvature. Five comments are made. First, the methods are not suitable for Problem \ref{problem1} as problem paths are needed that are not always and necessarily operating at the maximum curvature rate limit. Second and interestingly, in general, none of the 9 cases can handle Problem \ref{problem2} (e.g. for transitions as shown in Fig. \ref{fig_prob2_3R}). The \verb!Minimal longitudinal width!-case resembles a transition as shown in Fig. \ref{fig_prob2_3R}, but it has a specific condition $d=R_\text{big}$ (Sect. 2.3.7 in \cite{sabelhaus2013using}) that limits its usage. Another reason is that \cite{sabelhaus2013using} focussed only on lane-to-lane transitions and did not dicuss headland-to-lane transitions. Third, \cite{sabelhaus2013using} acknowledge that 'manoeuvres with start or end curvature $\kappa\neq 0$ are not feasible with continuous curvature turns'. In contrast, the proposed LP-based method this case can be handled, by adding an initial input constraint $\delta_0=\delta_\text{start}$ or $\delta_{N-1}=\delta_\text{end}$ to LP \eqref{eq_LP1} or \eqref{eq_LP2} for Problem \ref{problem1} or \ref{problem2}, respectively, and by observing that curvature can be related to steering via $\kappa=\frac{\text{tan}(\delta)}{l}$ (for the assumption of no-slip conditions of the wheels and instantaneous curvature of the rear axle). This is made possible as the LP-based approach incorporated state and input constraints in a structured manner. Fourth, as discussed for \cite{oh2024sharable}, the approach of \cite{sabelhaus2013using} is limited to a specific vehicle model (no slip condition). 

\begin{figure}
\centering
\includegraphics[width=6cm]{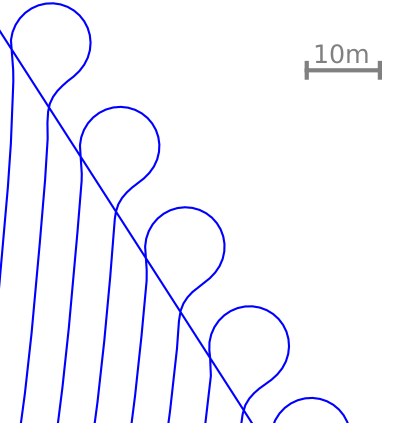}
\caption{Extending discussion. Proposed method can also be used for direct transitions between pairs of mainfield lanes. In this illustration, for non-symmetric mainfield lane start and end positions, pear-shaped steering maneuevers naturally result.}
\label{fig_f12}
\end{figure}

The proposed spatial LP-method can also handle lane-to-lane transitions. For illustration see Fig. \ref{fig_f12} with multiple non-symmetric start and end points and an interlane-distance of 5 m. The simplicity of the proposed method is that for the lane-to-lane transitions a Dubins-path is computed between the intersection points of headland path and mainfield lanes. The Dubins path is then used as the reference path in the second hierarchical step solving a spatial-based LP to obtain a refined path. This underlines the flexibility of the proposed method, in particular for shaping solutions to desired behavior.

A brief comment about path planning methods by neural networks is made (see a simplistic algorithm in \cite{graf2020automating} and the references therein). The key argument against the employment of neural networks is that both path planning Problems \ref{problem1} and \ref{problem2} deal with an \emph{infinite} number of possible tasks for headland path edges and headland-to-mainfield lane transitions and require a continuous action space. To encode a path planning module in a neural network, an enormous number of training examples \emph{and} their ground-truth solutions would be required, since there is an enormous variety in nature of different field contours. This would make a large training set mandatory. To underline the continous control aspect, see \cite{plappert2018parameter} where policies for continuous control are learned. However, each of the evaluation problems receives a reward either (i) upon \emph{moving forward} (\verb!HalfCheetah!, \verb!Hopper!, \verb!Swimmer!, \verb!Walker2D!, \verb!SparseHalfCheetah!, \verb!SwimmerGather!), or (ii) upon reaching a \emph{quasi-stationary} goal position \newline(\verb!InvertedDoublePendulum!, \verb!InvertedPendulum!, \verb!Reacher!, \newline\verb!SparseCartpoleSwingup!, \verb!SparseMountainCar!, \verb!SparseDoublePendulum!). \newline There is no option for path shaping by accounting for precision constraints. Training time and hardware requirements are not stated. In contrast, the proposed method here is based on linear programming, with a hyperparameter-free objective function, fast solution times, and can incorporate multivariate nonlinear vehicle dynamics and constraints on both states and control inputs in a structured manner to allow flexibility. Therefore, the proposed method can generate deterministic precision paths, which are required for both Problems \ref{problem1} and \ref{problem2}.

\cite{peng2023optimization} proposed an optimisation‐based algorithm in the time domain for headland turning under constraints imposed by headland geometry and obstacles, especially in orchard-like environments. In this paper, obstacles in the headland are not assumed, however, since obstacles are a relevant problem, the topic of obstacle avoidance within the spatial domain is briefly addressed. Based on previous work, three options have been proposed: (i) using a geometric path planner to solve the combinatorial corridor planning problem (\cite{plessen2017spatial}), (ii) using a linearisation-approach to model obstacles in the spatial domain (\cite{plessen2017trajectory}), and (iii) using path corridor heuristics (\cite{plessen2017trajectorytwo}). The last reference further discusses how velocity bounds within friction limits could be incorporated into spatial-based optimisation problems.

Finally, a comment about manual instead of autonomous driving is made. A visualisation, such as Fig. \ref{fig_f8}, gives a farmer a reference path plan and practical guidance for how to steer. The importance of generated path plans for manual driving should not be underestimated, since the decision about how to drive transitions between a headland path and mainfield lanes is often not straightforward in practice.

\section{Conclusion\label{sec_conclusion}}

This paper contributed to the task of path planning within agricultural fields by proposing a method based on a spatial domain transformation of vehicle dynamics and linear programming to address the two tasks of smoothing of headland path edges and headland-to-mainfield lane transitions. Within the agricultural context this is the first known approach suggesting to use a spatial domain vehicle dynamics representation in combination with a numerical optimisation approach to compute precision paths. The proposed method is of interest for both autonomous robot applications as well as as to provide high-level reference path plans for manual driving.

The main avenue of future work is the development of alternative heuristics to the 5-point piecewise-affine reference generation step for Problem Type \ref{problem1}. A second topic is the extension of the methods to also permit backward-motion in the path planning. This would permit three-point steering maneuvers involving both forward and backward motion at the transitions between headland path and mainfield lanes or headland path edges.

\bibliographystyle{model5-names} 
\bibliography{mybibfile.bib}
\nocite{*}

\end{document}